\documentclass[runningheads]{llncs}

\usepackage{eccv}

\usepackage{eccvabbrv}

\usepackage{graphicx}
\usepackage{booktabs}
\usepackage[accsupp]{axessibility}  %

\usepackage{hyperref}

\usepackage{orcidlink}
\usepackage{booktabs}       %
\usepackage{multirow}
\usepackage{amsmath,amssymb}
\usepackage{color}
\usepackage{caption}
\usepackage{subcaption}
\usepackage{graphicx}
\usepackage{xspace}
\usepackage{bbm}
\usepackage{makecell}
\usepackage{nccmath}
\usepackage{xcolor}
\usepackage{csquotes}
\usepackage{pifont}%
\usepackage{fontawesome}

\usepackage[ruled, lined, longend, linesnumbered]{algorithm2e}
\usepackage[capitalize]{cleveref}

\newcommand{\mc}{\mathcal}
\newcommand{\R}{\mathbb{R}}
\newcommand{\bS}{\mathbb{S}}

\newcommand{\D}{\mathcal{D}}
\newcommand{\x}{\boldsymbol{x}}

\newcommand{\norm}[1]{\left|\left|#1\right|\right|}

\makeatletter
\DeclareRobustCommand\onedot{\futurelet\@let@token\@onedot}

\def\@onedot{\ifx\@let@token.\else.\null\fi\xspace}

\def\eg{\emph{e.g}\onedot} 
\def\ie{\emph{i.e}\onedot}

\def\etal{\emph{et al}\onedot}

\crefname{section}{Sec.}{Secs.}
\Crefname{section}{Section}{Sections}
\Crefname{table}{Table}{Tables}
\crefname{table}{Tab.}{Tabs.}
\crefname{figure}{Fig.}{Figs.}

\definecolor{myGreen}{rgb}{0,0.5,0.3}
\definecolor{myRed}{rgb}{0.6,0,0}

\begin{document}

\title{SynCDR : Training Cross Domain Retrieval Models with Synthetic Data} 

\titlerunning{SynCDR : Training Cross Domain Retrieval Models with Synthetic Data}

\author{
Samarth Mishra$^{1}$ 
\quad Carlos D. Castillo$^{2}$ \quad Hongcheng Wang$^{2}$ \\ 
Kate Saenko$^{1,3}$ \quad Venkatesh Saligrama$^{1}$ \\
}

\authorrunning{S.~Mishra et al.}

\institute{$^1$Boston University \quad $^2$Amazon 
\quad $^3$Meta AI (FAIR)
}

\maketitle

\begin{abstract}
\vspace{-5mm}
   In cross-domain retrieval, a model is required to identify images from the same semantic category across two visual domains. For instance, given a sketch of an object, a model needs to retrieve a real image of it from an online store's catalog. A standard approach for such a problem is learning a feature space of images where Euclidean distances reflect similarity. Even without human annotations, which may be expensive to acquire, prior methods function reasonably well using unlabeled images for training. Our problem constraint takes this further to scenarios where the two domains do not necessarily share any common categories in training data. One instance where this can occur is when the two domains in question come from different versions of some biometric sensor recording identities of different people. We posit a simple solution, which is to generate synthetic data to fill in these missing category examples across domains. This, we do via category preserving translation of images from one visual domain to another. We compare approaches specifically trained for this translation for a pair of domains, as well as those that can use large-scale pre-trained text-to-image diffusion models via prompts, and find that the latter can generate better replacement synthetic data, leading to more accurate cross-domain. Our best SynCDR model can outperform prior art by up to 15\%. Code for our work is available at \href{https://github.com/samarth4149/SynCDR}{https://github.com/samarth4149/SynCDR} 
   
\end{abstract}
\section{Introduction} \label{sec:intro}

\begin{figure*}[]
    \centering
    \includegraphics[width=\linewidth,trim=0cm 0cm 0cm 0cm,clip]{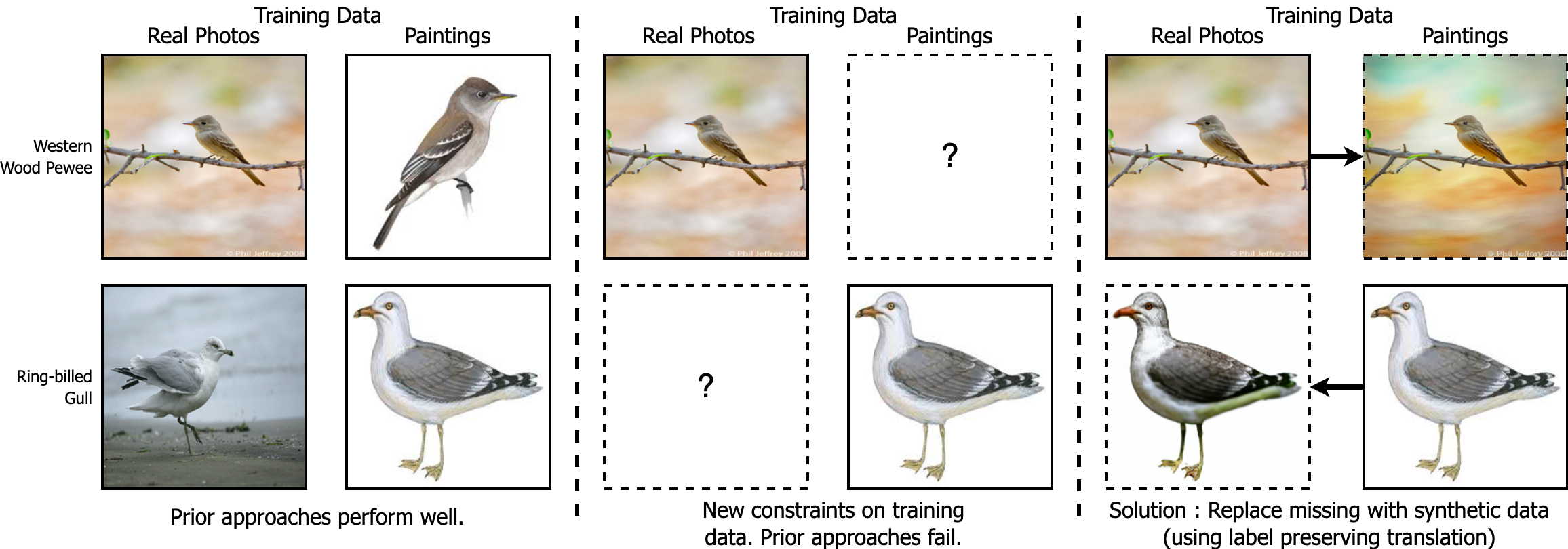}
    \caption{{\small 
    \textbf{Motivation.} Cross Domain Retrieval problems show up in many applications, and prior work has developed solutions in different scenarios, including when labeled data is absent~\cite{kim2021cds}. What these solutions rely on however is the presence of same category data in both domains so similar pairs can be discovered. When these are missing, such approaches can fail. Our solution is to make up for these missing examples using synthetic data (generated via label preserving translation). While such data may not be a perfect replacement (\eg the real image generated from the painting in the above example may not be entirely realistic because of a white background), we show that they are still useful for training cross-domain retrieval models.}}
    \label{fig:fig1} 
    \vspace{-6mm}
\end{figure*}

It is a non-trivial task for deep recognition models to generalize across visual domains, \ie, a model trained to recognize objects only in charcoal sketches fails to do so in real images due to a large training-test distribution shift~\cite{saenko2010adapting}. This is one key challenge in \textbf{cross-domain retrieval}, where one may desire to retrieve an item from an online catalog based on a hand-drawn sketch.

Albeit with plentiful human annotated data in both domains available for training, this problem is simpler, one cannot always rely on their presence due to the expense of curating them. Kim~\etal~\cite{kim2021cds} attempted to solve this problem with access to only unlabeled images from the two domains at training time. They used these to train an embedding model that could be used to retrieve similar images (determined by the category of the primary object in the image) across domains. Their training method of cross-domain self-supervision (CDS), relied on optimizing for contrastive criteria---one within and one across domains. While within domains, the criterion used was instance discrimination, across domains, it was entropy minimization. This meant that the network discovered similar pairs in training instances across domains and was trained to reinforce the similarity measure between such highly similar instances. It is important to note that how effective this optimization is depends on (a) the network being initialized with pre-trained weights that can result in some meaningful initial similarity measures and (b) the presence of semantically similar (or same category) examples across domains. 

While (a) may be available via the range of open-source pre-trained vision models (\eg \cite{rw2019timm, ilharco_gabriel_2021_5143773}), (b) is problem specific and may not always be available. For instance, in some biometric data collection process, if a sensor is upgraded at some point, identities recorded with the old sensor would not be available with the new one and vice-versa. See \cref{fig:fig1} for an example in a dataset of bird images~\cite{wah2011caltech}, where categories are defined by species of birds. In such scenarios with missing cross-domain pairs with same labels, we show that CDS is unable to learn representations that encode category semantics effectively across domains.

The solution we propose is to use synthetic data to fill in these gaps left by missing category examples in either domain. This we attempt to obtain via label-preserving translation across domains. Given an image in domain A, we wish to synthesize an image in domain B, which has the same category label as the original image. Note that this is done without access to the ground truth label information. An example is shown in \cref{fig:fig1}. We then train our cross-domain retrieval model with synthetic data (SynCDR) using self-supervision losses used by prior work~\cite{kim2021cds} as well as a new loss based on pseudo positive pairs, that utilizes these pair labels between a real image and its synthetic counterpart in the other domain, that we get for free from label-preserving cross-domain translations. 

How should we generate the aforementioned synthetic data? One strategy could be to use image-to-image translation trained on unpaired data. While there is a gamut of prior work here~\cite{zhu2017unpaired, isola2017image, huang2017arbitrary, huang2018munit, huang2018multimodal, lee2018diverse, chang2020domain, park2020contrastive}, their focus has primarily been on perceptual quality of generated images rather than their use as synthetic data for training recognition models. For instance, Lee~\etal~\cite{lee2018diverse} used their GAN based translation method only for MNIST$\leftrightarrow$MNIST-M digit recognition, which, prior work from the same year~\cite{shu2018dirt} reported better results on without using any synthetic data. In our experiments, we utilize contrastive unpaired translation (CUT)~\cite{park2020contrastive} as a representative approach, which is first trained on unlabeled data from the two domains for translation. This process is expensive, and can be avoided by using other pre-trained models in a text-guided fashion as we elucidate next. 

The recent advent of large pre-trained text-to-image diffusion models such as Stable Diffusion~\cite{rombach2022high}, has facilitated text-guided image editing. This, we can also use for image translation without any subsequent target-specific training, using approaches like SDEdit~\cite{meng2021sdedit} and InstructPix2Pix~\cite{brooks2023instructpix2pix}. Because by design, these methods attempt to closely mimic the input image, we find that they often fail to accurately represent the target domain. Alternatively, they can edit the image too much (based on a control parameter) and not preserve category labels. A better solution, we found, is using a personalization approach such as ELITE~\cite{wei2023elite}, that can learn to encode input image content into Stable Diffusion's vocabulary and more naturally combine it with the target domain, thus generating better synthetic data (see \cref{fig:syn_examples_motivation} for examples).

\noindent Our contributions can be summarized as follows: 
\begin{itemize}
    \item We show prior work is ineffective for training cross-domain retrieval models when same category pairs are not available across domains. 
    \item We make up for these missing examples with synthetic data using label-preserving translation across domains and develop a loss to use pseudo positive pair labels that we get for free from such translations for training.
    \item We compare multiple synthetic data generators, including translation methods trained specifically on a pair of domains, as well as those leveraging large-scale pre-trained generative models, which can be used without any target-specific training, via the use of prompts.
    \item With the best synthetic data, SynCDR realizes performance benefits of $\sim$15\% on the DomainNet dataset~\cite{peng2019moment} and $\sim$6\% on a challenging fine-grained recognition task over images of different bird species~\cite{wah2011caltech, wang2020progressive}, compared to baselines that do not use any synthetic data.
\end{itemize}

\section{Related Work}

\noindent \textbf{Cross-domain Image Retrieval.} Given a ``query'' image, there are multiple practical scenarios where one may expect to fetch similar images from a database. Often, there might be a shift in visual domains between the query image and the image database. Examples include retrieving person images from face sketches, retrieving product images using sketches/cell phone pictures, among others. Given these uses, there has been an array of prior research with multiple summary review articles~\cite{smeulders2000content, datta2008image}. In the deep learning era, solutions for this problem involve learning a feature space where euclidean distances can reflect similarity and this metric is used to rank ``target'' images in the database, retrieving the top-1 or top few. Some characteristic approaches here have used category label information~\cite{sangkloy2016sketchy}, automatically extracted attribute annotations~\cite{huang2015cross}, or triplet annotations~\cite{yu2016sketch, song2017deep} for training a deep feature extractor. 

\noindent \textbf{Self-supervised Feature Learning for Cross-Domain Retrieval.} While approaches above rely on labels for training, these may be expensive in labor-hours to collect. In the absence of any labels representation learning can be done by optimizing for different self-supervision objectives~\cite{gidaris2018unsupervised, chen2020simple, he2020momentum, chen2021exploring}. Kim~\etal~\cite{kim2021cds}, followed by others~\cite{yue2021prototypical, hu2022feature}, demonstrated that category discriminative domain-adaptive features can be learned without the presence of any labels using self-supervised criteria within and across domains. This, however, relies on the presence of same categories in both domains in training data. If unavailable, the cross-domain self-supervision losses can degrade retrieval performance by training the model to match unrelated instances. This is the problem we tackle in this paper, by using synthetically generated examples to make up for the missing categories of data.

\noindent \textbf{Unpaired Image-to-Image Translation.} One primary approach we adopt is translating an image such that its \emph{content} is preserved but the style mimics the alternate domain. This way, we have synthetically added examples for missing categories in either domain. While there are prior work that can perform this translation when trained on paired images~\cite{isola2017image}, we do not have such paired data. What we can use are methods that can be trained for this task with unpaired data from two domains~\cite{zhu2017unpaired, huang2018multimodal, lee2018diverse, park2020contrastive}. We use CUT~\cite{park2020contrastive} as a representative approach in our experiments.

\begin{figure*}[h!]
    \centering
    \includegraphics[width=0.90\linewidth,trim=0cm 0cm 0cm 0cm,clip]{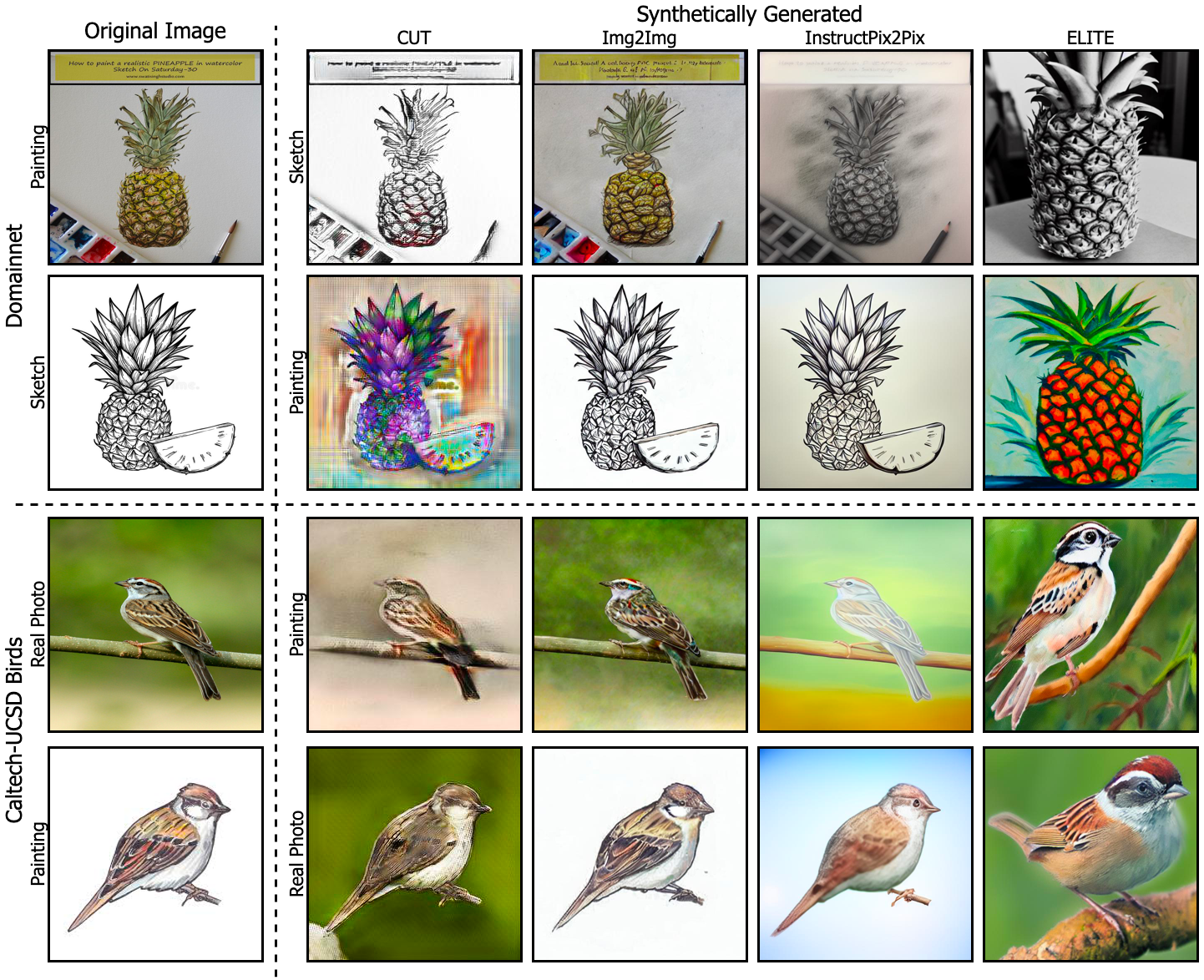}
    \caption{{\small \textbf{Synthetic examples from different translation methods.}  We compare synthetic data generated using 4 methods. Since ELITE is not specifically restricted to closely mimicking the original image, we can generate more natural examples of target domain data, which serves as better synthetic replacement for missing real data. For more discussion, refer to \cref{subsec:syn_methods}.}}
    \label{fig:syn_examples_motivation} 
    \vspace{-6mm}
\end{figure*}

\noindent \textbf{Large Scale Pre-trained Text-to-Image Generative Models.} The quality and diversity of generated images from diffusion models has given rise to models that can follow text instruction in image generation, via training on large-scale image-text paired data~\cite{ramesh2022hierarchical, saharia2022photorealistic, rombach2022high}. These models have enabled several further applications---image editing~\cite{meng2021sdedit, hertz2022prompt, brooks2023instructpix2pix}, personalization~\cite{gal2022image,ruiz2023dreambooth,wei2023elite}, text-to-3D generation~\cite{poole2022dreamfusion}, to name a few. Using the aforementioned image editing approaches~\cite{meng2021sdedit, brooks2023instructpix2pix} can allow us to perform cross-domain translations for our use-case, as we shall see in \cref{sec:method}. These methods, by design, attempt to closely resemble the input image, allowing control over the extent of resemblance via a single parameter. We find in many cases, they either cannot preserve image content well, or do not result in a large enough edit to effectively represent the target domain (see an example of the latter in \cref{fig:syn_examples_motivation}). Using a personalization approach like ELITE~\cite{wei2023elite} this can be mitigated. ELITE can encode a given image's content as a new token in the vocabulary of Stable Diffusion, and then combine this more naturally with the target domain, without the restriction of closely resembling the given source image.

\noindent \textbf{Synthetic Data from Text-to-Image Diffusion Models.} Given the high fidelity and level of control that models like Stable-Diffusion~\cite{rombach2022high} and Imagen~\cite{saharia2022photorealistic} can provide, recent work has evaluated their use as data sources for training recognition models. \cite{sariyildiz2022fake} and \cite{azizi2023synthetic} used them for generating Imagenet-like synthetic data and studied scaling and transfer properties of recognition models trained on them. Tian~\etal developed a method StableRep~\cite{tian2023stablerep}, that used multiple generations from a single caption in a multi-positive contrastive loss to train an image representation with 20M synthetic images, that is as accurate as CLIP~\cite{radford2021learning} trained on 50M real images. \cite{trabucco2023effective, dunlap2023diversify, qraitem2023fake} used synthetic data from these models in classification problems with few/biased training examples. We extend this general research topic with our approach using these models for generating cross-domain translations.

\section{Approach} \label{sec:method}

\noindent\textbf{Notation.} Let the two domains be A and B. At training time,   two sets of images are available (with no accompanying labels), $\D_A = \{\x^{A}_i\}_{i=1}^{n_A}$ and $\D_B = \{\x^{B}_i\}_{i=1}^{n_B}$. The ground truth category labels are unknown but let each image in $\D_A$ come from a category label in $Y_A$, and each image in $\D_B$ come from a category label in $Y_B$. It is known that $Y_A \cap Y_B = \phi$ (This is not a strict requirement for SynCDR to function and we report results where $Y_A$ and $Y_B$ share some categories in \cref{subsec:cat_overlap}). The outcome of training is a feature extractor $f : \mc{X} \rightarrow \bS^{d}$ (where $\mc{X}$ is the support of all images, and $\bS^{d}$ the $d$-dimensional unit hypersphere $\bS^{d} = \{\x; \x \in \R^{d}, \norm{x}_2 = 1\}$), such that cosine similarities in the output feature space reflect category semantics. 

For evaluation are provided two sets $\D^{(test)}_A$ and $\D^{(test)}_B$ both of which have images with known category labels and each set has at least one image from each label in $Y_A \cup Y_B$. At test time, given an image $\x^{A}$ in $\D^{(test)}_A$, it is desired to fetch images of the same category label from $\D^{(test)}_B$, and vice-versa given an image in domain B. This is done by ranking in decreasing order $\{f(\x^{A})^\top f(\x^{B}) ; \x^{B} \in \D^{(test)}_B \}$ and picking top K to calculate precision@K (for different values of K).

\noindent\textbf{Synthetic Data.} For our solution we use synthetic data generators $G_{A \rightarrow B}$, which, given an image $\x^{A} \in \mc{X}^{A}$ ($\mc{X}^{A}$ being the support of images from domain A), can generate a synthetic image $\x_{syn}^{B}$ with the content of $\x^A$ in the style of domain $B$. We also have $G_{B \rightarrow A}$, that can do this in the other direction. Using these, we generate $\D_{syn, B} = \{G_{A \rightarrow B} (\x) ; \x \in \D_A \}$ and $\D_{syn, A} = \{G_{B \rightarrow A} (\x) ; \x \in \D_B \}$. Per our described motivation (\cref{fig:fig1}), we can now combine these synthetic data with real data to make up for missing categories and train with a prior approach such as CDS~\cite{kim2021cds} on this entire set. However, the manner in which we generated synthetic data provides us more information to use as described next.

\noindent\textbf{Pseudo Positive Pairs (PPP).} Given that $G_{A \rightarrow B}$ preserves content for image $\x^{A}$, we can use the fact that $\x^{A}$ and $\x_{syn}^{B}$ are the same semantic category to train our model. In this context, we refer to $\x^{A}$ and $\x_{syn}^{B}$ as pseudo-positive pairs (PPP) and use them as follows in a PPP loss.  

During training, we sample batches of size $m$ from real domain-A data $X_A = [\x^{A}_i]_{i=1}^{m}$ and $X_{syn, B} = [\x^{B}_{syn, i}]_{i=1}^m$. Using the feature extractor $f$, we define the loss

\begin{align} \label{eq:ppp_loss}
    L_{PPP} (X_A, X_{syn, B}) &= \frac{1}{m} \sum_{\x^{A} \in X_A} -\log\left(\frac{\exp(f(\x^A)^\top f(\x_{syn}^B) )}{\sum_{\x \in X_{syn, B}}\exp(f(\x^A)^\top f(\x) ) } \right) \nonumber\\
    &+ \frac{1}{m} \sum_{\x_{syn}^{B} \in X_{syn, B}} -\log\left(\frac{\exp(f(\x_{syn}^B)^\top f(\x^A) )}{\sum_{\x \in X_A}\exp(f(\x_{syn}^B)^\top f(\x) ) } \right)
\end{align}

which is a contrastive loss matching the real example $\x^{A}$ to its synthetic counterpart $\x_{syn}^{B} = G_{A \rightarrow B}(\x^{A})$ in a batch of synthetic examples and likewise in the opposite direction for a synthetic example, matching it to its real counterpart in a batch of real examples.

\noindent\textbf{Training SynCDR.} For training we combine $L_{PPP}$ with the CDS criterion~\cite{kim2021cds} ($L_{CDS}$), and minimize the following loss via minibatch stochastic gradient descent. To meet length constraints, we excluded a precise definition of $L_{CDS}$ and refer readers to Sec. 3 and Fig. 3 of \cite{kim2021cds}. For the sake of current discussion, we simply describe it as a combination of in-domain and cross-domain contrastive criteria given two sets of unlabeled images from two domains.
\begin{align} \label{eq:total_loss}
    L &= L_{CDS} (\D_A \cup \D_{syn, A}, \D_B \cup \D_{syn, B}) \nonumber \\
    &+ \frac{1}{2}\left[ L_{PPP} (\D_{A}, \D_{syn, B}) + L_{PPP} (\D_{B}, \D_{syn, A}) \right]
\end{align}
where the two arguments of $L_{CDS}$ reflect the unlabeled datasets in the two domains and we abuse notation described in \cref{eq:ppp_loss} to describe the loss for entire sets rather than minibatches.

\vspace{-3mm}
\subsection{Methods for Synthetic Data Generation} \label{subsec:syn_methods}
We use the following methods for generating our synthetic cross-domain translations. In \cref{fig:syn_examples_motivation}, we provide examples for translations generated by each.

\noindent\textbf{Contrastive Unpaired Translation (CUT)~\cite{park2020contrastive}} is an approach that trains GAN image generators for the requisite translation task using unpaired images from the two domains. For training this generator, we use all images in $\D_A$ and $\D_B$. The training process is quite expensive, requiring $\sim$20 hrs on a single NVIDIA Tesla V100 gpu.

\noindent\textbf{Img2Img/SDEdit~\cite{meng2021sdedit}} uses a pre-trained Stable Diffusion model for image editing by partially following the diffusion forward noising process (so that not all content in the input image is lost due to noising) and then denoising based on a text prompt. For instance in the example in \cref{fig:syn_examples_motivation}, for translating real images of birds from the CUB dataset to paintings, we used the prompt ``A painting of a bird.'' and for translating sketches to paintings for Domainnet objects, we used ``A painting of an object.''. The method allows for specification of \emph{edit strength} $\in [0, 1]$, which is the fraction of diffusion noising steps to follow. The value $0$ leads to no edit and $1$ leads to completely following the prompt and retaining none of the input image information. In our experiments, we set this parameter to $0.3$ based on validation performance. For prompts used for all datasets and other generation parameters, we refer readers to \cref{sec:syn_details}.

\noindent\textbf{InstructPix2Pix~\cite{brooks2023instructpix2pix}} is a Stable-Diffusion model fine-tuned and adapted for following natural language instructions for editing images. For converting to paintings (for both photos of birds as well as sketches of objects from Domainnet), we input the original image and the prompt ``Convert to a painting.''. With InstructPix2Pix, we can provide a parameter $image\_guidance\_scale \ge 1$. The larger the value of this parameter, more is the output image similar to the input image. In our experiments, we set this parameter to $1.2$ based on validation performance. Full prompts and other generation parameters for all datasets and additional examples are in \cref{sec:syn_details}.

The above translation methods attempt to closely resemble the original image in translations and we find that this is often not enough to accurately represent data in the other domain. In such scenarios, images may need additional content to make it look like the target domain. For instance, in \cref{fig:syn_examples_motivation}, while the above methods do a decent job of converting paintings to pencil sketches or real photos to paintings, the translation in the opposite direction is much poorer. In translating a sketch to a painting (row 2), Img2Img and InstructPix2Pix fail to add color and CUT adds some incorrect colors based on what it could learn from unpaired images. Similar is the case in converting a painting of a bird to a realistic photograph (row 4), where a background may be expected but above methods fail to generate one (CUT does generate one, but does not generate an natural looking photograph). While Img2Img and InstructPix2Pix allow for increasing the amount of edit via a controllable parameter, in \cref{sec:syn_details}, we show that this often leads to non-preservation of image category. The following generation method can avoid these issues.

\noindent\textbf{ELITE~\cite{wei2023elite}} is a personalization approach based on Stable Diffusion and a CLIP-based image encoder. The encoder encodes an object instance from an image as a word in Stable Diffusion's vocabulary, which can then be used for generation with prompts. In the example in \cref{fig:syn_examples_motivation}, for generating a painting (from either a real photo or a sketch), we input the image to ELITE's image encoder and use the prompt ``A painting of $\langle S \rangle$.'' (where $\langle S \rangle$ is the image encoded as a word, or more precisely, a token embedding). For full details, refer to \cref{sec:syn_details}. 

Instead of being restricted by having to closely resemble the input image, ELITE can learn the object's properties and more naturally combine them with those of the target domain. In \cref{fig:syn_examples_motivation} (rows 2 and 4), we see that this results in the most natural looking target domain images with the object from the original image. As we shall see in \cref{sec:expts}, this serves as the most effective synthetic data for learning cross-domain retrieval models.

\vspace{-3mm}
\section{Experiments} \label{sec:expts}
\vspace{-1mm}
\noindent\textbf{Datasets and Evaluation.} We ran experiments on three different datasets : DomainNet~\cite{peng2019moment}, CUB~\cite{wah2011caltech, wang2020progressive} and Office-Home~\cite{venkateswara2017deep}. We experimented with the 126 class subset of DomainNet used in \cite{kim2021cds} with 3 domains : clipart, painting and sketch. For CUB, we use two domains : Real (coming from the CUB-200-2011 data~\cite{wah2011caltech}) and Paintings (coming from \cite{wang2020progressive}). In CUB, we removed categories that had fewer than 5 paintings, leaving us with 187 categories. The Office-Home dataset has 65 categories in 4 domains : Art, Clipart, Product and Real World. 

We experimented with each possible pair of domains in each dataset. For the experiments, we made a 50:20:30 split of the data in the two domains using the first for training, second for validation and third for testing. For each pair of domains we split the training data into two disjoint sets of categories (of equal size) and run two experiments : one where category set 1 is from domain A and category set 2 is from domain B, and the other where this is reversed. Note that testing in either of these cases takes place on the same data, and we report average Prec@1 of retrieval in either direction (given domain A image, fetch a domain B image of same category and vice-versa), and the overall average of these over all pairs of domains in the dataset. In \cref{subsec:full_ret_results}, we also report Prec@5 and Prec@15.  

\noindent\textbf{Other Implementation Details.} Similar to CDS~\cite{kim2021cds}, we used a ResNet-50~\cite{he2016deep} backbone pre-trained on Imagenet followed by a fully connected layer and $L_2$ normalization as our feature extractor. We trained for 15 epochs and perform early stopping based on average validation Prec@1. Other training hyperparameters are kept the same as CDS~\cite{kim2021cds}. In our approach using synthetic data, we generated one synthetic example per real example in the training set. For full training details, please refer to \cref{sec:train_details}. 

\vspace{-3mm}
\subsection{Cross Domain Retrieval Results} \label{subsec:cdr_results}

\begin{table*}[t]
\begin{center}
\caption{\textbf{Cross Domain Retrieval Prec@1 in different scenarios of DomainNet.} As highlighted in \cref{fig:fig1}, prior art (CDS) cannot perform well (and is only about as good as a pre-trained ImageNet model) in this scenario where training data does not have examples from the same categories across domains. The cross-domain criterion from CDS, impedes performance and using only the In-domain ID criterion for training performs better. As seen in the results, synthetic data can make up for missing real examples and lead to improved performance. ELITE provides the best quality synthetic data for SynCDR, resulting in best cross-domain retrieval performance.
}
\vspace{-2mm}
\label{tab:domainnet_results}
\scalebox{0.75}{
\begin{tabular}{@{}llcccc@{}}
\toprule
\textbf{Model} & \textbf{Synthetic Data} & \textbf{Clipart - Painting} & \textbf{Clipart - Sketch} & \textbf{Painting -   Sketch} & \textbf{Average} \\
\midrule
ImageNet (pt) & \multicolumn{1}{c}{-} & 27.4 & 24.6 & 29.2 & 27.1 \\
CDS~\cite{kim2021cds} & \multicolumn{1}{c}{-} & 28.6 ± 0.7 & 24.5 ± 0.6 & 30.5 ± 0.6 & 27.8 ± 0.6 \\
In-domain ID~\cite{wu2018unsupervised} & \multicolumn{1}{c}{-} & 30.7 ± 0.4 & 27.5 ± 0.2 & 31.8 ± 0.4 & 30.0 ± 0.4 \\
\midrule
\multirow{4}{*}{SynCDR (Ours)} & CUT & 38.8 ± 0.8 & 40.1 ± 0.4 & 44.7 ± 0.6 & 41.2 ± 0.6 \\
 & Img2Img & 38.2 ± 0.7 & 34.5 ± 0.7 & 37.4 ± 0.5 & 36.7 ± 0.6 \\
 & InstructPix2Pix & 41.6 ± 0.6 & 39.8 ± 0.7 & 42.5 ± 0.7 & 41.3 ± 0.7 \\
 & ELITE & \textbf{45.4 ± 0.7} & \textbf{44.2 ± 0.4} & \textbf{46.4 ± 1.3} & \textbf{45.3 ± 0.9} \\
 \bottomrule
\end{tabular}
}
\vspace{-5mm}
\end{center}
\end{table*}

\noindent\textbf{Methods for Comparison.} We compare 4 versions of SynCDR using different synthetic data generation methods. We also compare them to prior work using no synthetic data. CDS is the method proposed by Kim~\etal~\cite{kim2021cds} which uses a combination of in-domain instance discrimination~\cite{wu2018unsupervised} (ID) and cross-domain matching. As another comparison, we drop the latter of these two criteria and train only with in-domain ID loss. As we shall see, in the absence of same category examples across domains, this can lead to better performance. Finally, we also report the performance of using simply an ImageNet-1K pre-trained ResNet-50 backbone, which serves as the initialization for all the other methods.

We digress briefly to note that while large image embedding models based on vision transformers~\cite{dosovitskiy2020image} and CLIP~\cite{radford2021learning} pre-training are available~\cite{ilharco_gabriel_2021_5143773} and provide robust image embeddings across domains, they may be expensive to store and run on smaller devices at inference time. For instance, a CLIP ViT-B model (the smallest CLIP available) has 160M parameters compared to a ResNet-50's 25M. We thus only use large pre-trained models at training time. Although knowledge distillation~\cite{hinton2015distilling} is possible from these large CLIP models to smaller ones, as shown by Sun~\etal~\cite{sun2023dime} (in Fig. 3 of their paper), it still requires a large number of images and text captions (a few tens of millions). In \cref{subsec:distill}, we show that distillation is not effective with the limited size of our training datasets (which also have no captions available).

Each of \cref{tab:domainnet_results,,tab:cub_results,,tab:office_home_results} reports the results of cross-domain retrieval (Prec@1) across all pairs of domains for a given dataset. Each cell (except the ones for the pre-trained models) reports mean and standard deviation over 3 different training runs. The ``Average'' column reports the mean and the pooled standard deviation over the other columns in the table. We additionally report Prec@5 and Prec@15 in \cref{subsec:full_ret_results}, while noting that they follow the same trends as seen in Prec@1.

\begin{table}[t]
\begin{center}
\caption{\textbf{Cross Domain Retrieval Prec@1 for CUB$\leftrightarrow$CUB-Paintings.} Compared to the Imagenet pre-trained model which serves as initialization for all methods, we find both CDS and In-domain ID help improve performance by a small amount. ELITE's synthetic data again leads to the best performance with SynCDR. (For more discussion, refer to \cref{subsec:cdr_results})}
\vspace{-2mm}
\scalebox{0.9}{
\begin{tabular}{@{}llc@{}}
\toprule
\textbf{Model} & \textbf{Synthetic Data} & \textbf{Painting - Real} \\
\midrule
ImageNet (pt) & \multicolumn{1}{c}{-} & 20.5 \\
CDS~\cite{kim2021cds} & \multicolumn{1}{c}{-} & 22.0 ± 0.8 \\
In-domain ID~\cite{wu2018unsupervised} & \multicolumn{1}{c}{-} & 21.4 ± 1.2 \\
\midrule
\multirow{4}{*}{SynCDR (Ours)} & CUT & 23.2 ± 0.5 \\
 & Img2Img & 22.0 ± 0.5 \\
 & InstructPix2Pix & 21.5 ± 0.7 \\
 & ELITE & \textbf{28.2 ± 0.4} \\
 \bottomrule
\end{tabular}
} %
\vspace{-4mm}
\label{tab:cub_results}
\end{center}
\end{table}

\begin{table*}[t]
\begin{center}
\caption{\textbf{Cross Domain Retrieval Prec@1 in different scenarios of Office-Home.}  
As with other benchmarks, we find that SynCDR with ELITE generated data results in best performance on average over all scenarios in Office-Home. We see synthetic data results in relatively small performance improvements in scenarios with small domain gaps (\eg Product-Real), but more significant improvements in larger domain gap scenarios (\eg those involving Clipart).
Refer to \cref{subsec:cdr_results} for further discussion.}
\vspace{-3mm}
\scalebox{0.55}{
\begin{tabular}{@{}llccccccc@{}}
\toprule
\textbf{Model}                 & \textbf{Synthetic Data} & \textbf{Art - Clipart} & \textbf{Art - Product} & \textbf{Art - Real} & \textbf{Clipart -   Product} & \textbf{Clipart - Real} & \textbf{Product - Real} & \textbf{Average}    \\ 
\midrule
ImageNet (pt)                  & \multicolumn{1}{c}{-}   & 34.5                   & 48.6                   & 54.6                & 42.2                         & 43.9                    & 68.6                    & 48.7                \\
CDS                            & \multicolumn{1}{c}{-}   & 36.5 ± 0.3             & 50.0 ± 0.4             & 54.9 ± 0.3          & 43.0 ± 0.2                   & 45.2 ± 0.3              & 69.1 ± 0.2              & 49.8 ± 0.3          \\
In-domain ID                             & \multicolumn{1}{c}{-}   & 36.3 ± 0.3             & 50.0 ± 0.4             & 54.7 ± 0.3          & 43.4 ± 0.4                   & 45.4 ± 0.4              & 69.0 ± 0.3              & 49.8 ± 0.3          \\
\midrule
\multirow{4}{*}{SynCDR (Ours)} & CUT                     & 38.3 ± 0.4             & 50.8 ± 0.2             & \textbf{55.5 ± 0.6} & 43.3 ± 0.5                   & 46.6 ± 0.3              & \textbf{69.3 ± 0.3}     & 50.6 ± 0.4          \\
                               & Img2Img                 & 36.4 ± 0.3             & 49.9 ± 0.3             & 54.9 ± 0.5          & 43.4 ± 0.4                   & 45.2 ± 0.4              & 69.0 ± 0.3              & 49.8 ± 0.4          \\
                               & InstructPix2Pix         & \textbf{38.6 ± 0.5}    & 50.2 ± 0.2             & 54.7 ± 0.3          & 43.9 ± 0.2                   & \textbf{46.7 ± 0.4}     & 69.0 ± 0.2              & 50.5 ± 0.3          \\
                               & ELITE                   & 38.3 ± 0.5             & \textbf{51.3 ± 0.3}    & 55.4 ± 0.3          & \textbf{44.1 ± 0.6}          & 46.3 ± 0.5              & \textbf{69.3 ± 0.3}     & \textbf{50.8 ± 0.4} \\
\bottomrule
\end{tabular}
}

\label{tab:office_home_results}
\vspace{-6mm}
\end{center}
\end{table*}

\noindent\textbf{DomainNet.} \cref{tab:domainnet_results} reports the results. We can see that CDS and ID both improve on top of ImageNet pre-training but CDS, having the negative effect of incorrect matchings from the cross-domain loss, performs poorer. All SynCDR variants improve on this performance, with the worst being Img2Img. We find that the best performing synthetic data from Img2Img can only be attained with a fairly low edit strength. This is because using a higher edit strength substantially trades off Img2Img's capacity to preserve the image content.

We also find that CUT and InstructPix2Pix perform at par with each other on average, but each is better in different scenarios. InstructPix2Pix is poorer in scenarios involving sketch since it is worse at ``filling in color'' into pencil sketches (as seen from \cref{fig:syn_examples_motivation}), when simply prompted to convert a sketch to a painting or a clipart. On the other hand, CUT can add color to sketches based on it's training on the two domains, even though these colors may not always be correct. In the clipart-painting scenario, InstructPix2Pix has a bigger edge in the quality of data being able to rely on Stable Diffusion's prior understanding of these domains (more examples in \cref{sec:more_qual_examples}). Finally, we see that synthetic data from ELITE performs the best. Aside from generating good quality data, it does not suffer from InstructPix2Pix's limitations in adding color to sketches since ELITE parses the object into a textual token and then generates a painting of it, without having to be restricted by the amount of edit between the original and the generated image.

\noindent\textbf{CUB.} \cref{tab:cub_results} reports the results for CUB. Compared to Domainnet, this dataset involves a much more fine-grained recognition task. We find that both CDS and ID provide improved performance over ImageNet pre-training albeit by a small amount. CDS does not seem to be disadvantaged by its cross-domain matching criterion possibly because it finds images that are similar enough across different bird categories. We find that CUT results in better synthetic data and consequently improves performance more than Img2Img or InstructPix2Pix (also seen in \cref{fig:syn_examples_motivation}, row 4). This is possibly because most realistic photos of birds have a more consistent style than paintings of objects in DomainNet, where CUT is less accurate with adding colors (see \cref{fig:syn_examples_motivation} row 2). Again, similar to the case of DomainNet, ELITE from its ability of parsing objects and more naturally incorporating them in the target domain, leads to best quality synthetic data and hence, best retrieval performance.

\begin{table*}[t]
\begin{center}
\caption{\textbf{Ablating PPP --- Cross-domain retrieval results on DomainNet.} We see that using PPP, \ie, using the fact that cross-domain translations are label preserving for training, leads to big performance boosts for cross-domain retrieval. (up to 10\% on average with synthetic data generated with ELITE)}
\label{tab:ppp_ablate}
\scalebox{0.7}{
\begin{tabular}{@{}llcccc@{}}
\toprule
\textbf{Method   (Synthetic Data)} & \textbf{Loss} & \textbf{Clipart - Painting} & \textbf{Clipart - Sketch} & \textbf{Painting -   Sketch} & \textbf{Avg} \\
\midrule
\multirow{2}{*}{SynCDR (CUT)} & CDS & 36.1 ± 0.8 & 36.8 ± 1.1 & 35.6 ± 3.8 & 36.2 ± 2.4 \\
 & CDS + PPP & 38.8 ± 0.8 & 40.1 ± 0.4 & 44.7 ± 0.6 & 41.2 ± 0.6 \\
\multirow{2}{*}{SynCDR   (Img2Img)} & CDS & 38.1 ± 0.6 & 33.8 ± 0.6 & 37.4 ± 0.4 & 36.4 ± 0.5 \\
 & CDS + PPP & 38.2 ± 0.7 & 34.5 ± 0.7 & 37.4 ± 0.5 & 36.7 ± 0.6 \\
\multirow{2}{*}{SynCDR (InstructPix2Pix)} & CDS & 39.9 ± 0.6 & 37.5 ± 0.8 & 40.9 ± 0.7 & 39.4 ± 0.7 \\
 & CDS + PPP & 41.6 ± 0.6 & 39.8 ± 0.7 & 42.5 ± 0.7 & 41.3 ± 0.7 \\
\multirow{2}{*}{SynCDR   (ELITE)} & CDS & 38.2 ± 0.6 & 31.9 ± 0.9 & 36.1 ± 0.8 & 35.4 ± 0.7 \\
 & CDS + PPP & 45.4 ± 0.7 & 44.2 ± 0.4 & 46.4 ± 1.3 & 45.3 ± 0.9 \\
 \bottomrule
\end{tabular}
}
\vspace{-6mm}
\end{center}
\end{table*}

\noindent\textbf{Office-Home.} The results for this dataset are in \cref{tab:office_home_results}. As with other benchmarks, we found that SynCDR with ELITE's synthetic data resulted in best cross-domain retrieval performance overall. We find on some domains similar to ImageNet data (such as Product-Real), an ImageNet pre-trained backbone performs quite well, and CDS and in-domain ID cannot improve much on this given they are only self-supervised training objectives functioning with no dataset specific labels. Additionally, we find that on such small domain gap scenarios (Product has realistic images on a white background, and some training images were even found to occur in both Product and Real domains), performance improvement from synthetic data is relatively small. This is because the problem of missing examples as shown in \cref{fig:fig1} is not as big with small domain gaps. On the other hand, when the domain gap is larger (\eg when Clipart is involved) SynCDR helps improve performance more significantly.

\begin{table*}[t]
\begin{center}
\caption{\textbf{Analyzing Synthetic Data with CLIP features.} On all pairs of domains in the DomainNet dataset, using CLIP as an image featurizer, we try to answer three questions : (1) How much does a synthetic generation method edit a real image on average (Distance to Source); (2) how well does the translation preserve category labels (NCM Accuracy); and (3) how well does the synthetic data mimic the real data that it replaces (Similarity to Real Target). ELITE, which leads to the best performing synthetic data brings about large edits, preserves labels reasonably well and mimics real data the best, out of the 4 different methods compared.}
\label{tab:domainnet_clip_scores}
\scalebox{0.8}{
\begin{tabular}{@{}l|cccc|cccc|cccc@{}}
\toprule
\multirow{2}{*}{\textbf{Synthetic Data}} & \multicolumn{4}{c|}{\textbf{Distance   to Source}} & \multicolumn{4}{c|}{\textbf{NCM   Accuracy}} & \multicolumn{4}{c}{\textbf{Similarity   to Real Target}} \\
\cmidrule{2-13}
& \textbf{c-p} & \textbf{c-s} & \textbf{p-s} & \textbf{avg} & \textbf{c-p} & \textbf{c-s} & \textbf{p-s} & \textbf{avg} & \textbf{c-p} & \textbf{c-s} & \textbf{p-s} & \textbf{avg} \\
\midrule
\textbf{CUT} & 0.41 & 0.30 & 0.39 & 0.37 & 0.45 & 0.63 & 0.47 & 0.52 & 0.57 & 0.62 & 0.61 & 0.60 \\
\textbf{Img2Img} & 0.15 & 0.14 & 0.17 & 0.15 & \textbf{0.67} & \textbf{0.72} & 0.58 & \textbf{0.65} & 0.57 & 0.64 & 0.59 & 0.60 \\
\textbf{InstructPix2Pix} & 0.29 & 0.24 & 0.23 & 0.25 & 0.58 & 0.65 & \textbf{0.61} & 0.61 & 0.60 & \textbf{0.64} & 0.62 & 0.62 \\
\textbf{ELITE} & \textbf{0.47} & \textbf{0.48} & \textbf{0.47} & \textbf{0.47} & 0.59 & 0.53 & 0.56 & 0.56 & \textbf{0.67} & 0.64 & \textbf{0.68} & \textbf{0.66} \\
\bottomrule
\end{tabular}
}
\vspace{-6mm}
\end{center}
\end{table*}

\vspace{-3mm}
\subsection{Ablating PPP}
As described in \cref{sec:method}, we use the fact that our cross-domain translations preserve category labels in the form of the pseudo-positive pairs (PPP) loss. In \cref{tab:ppp_ablate} we report the performance of methods where we ignore these pairs and simply combine synthetic and real data, training with the CDS criterion. Comparing with methods using no synthetic data \cref{tab:domainnet_results}, we find that the addition of synthetic examples does improve performance even when PPP is not used. Additionally, using PPP provides a big performance advantage (up to 10\% on average in the case of ELITE). This differs for different synthetic data generators. In \cref{subsec:further_analyzing}, we find that this boost from PPP is often low in cases where the amount of edit made by the translation method is small. 

\vspace{-3mm}
\subsection{Further Analyzing Synthetic Data} \label{subsec:further_analyzing}

In this section, we use a CLIP image featurizer~\cite{ilharco_gabriel_2021_5143773}, $f_{CLIP} : \mc{X} \rightarrow \bS^{d}$ (following notation from \cref{sec:method}, recall that $\bS^{d}$ is the d-dimensional unit hypersphere), to quantitavely evaluate synthetic data from different generation sources (results in \cref{tab:domainnet_clip_scores} on the Domainnet dataset). Here, we slightly abuse notation in that $\D_A$ and $\D_B$ now have examples from all categories in $Y_A \cup Y_B$ (previously $\D_A$ and $\D_B$ did not share any categories). Note that this is done here only for evaluating synthetic data generated by each method.

\noindent\textbf{How different are synthetic images from the real counterparts they were generated from?} To find this, we computed \textbf{Distance to Source}, reported in \cref{tab:domainnet_clip_scores}. Specifically, for an image $\x^{A} \in \D_A$ and its synthetic counterpart $\x_{syn}^{B} = G_{A \rightarrow B (\x^{A})}$, we compute $1 - f_{CLIP}(\x^{A})^\top f_{CLIP}(\x_{syn}^{B})$ and compute the average over all examples in $\D_A$ (note that this quantity is proportional to the square of the distance between the vectors, given they lie on the unit hypersphere). Similarly we compute the average of these distances for the opposite direction of translation over examples in $\D_B$ and report the average of the two for each pair of domains in \cref{tab:domainnet_clip_scores}. 

We find the least edit amount from Img2Img data, followed by InstructPix2Pix and CUT, while ELITE generates images that are the most distinct from source since it is not tied down to closely mimicking the source real image. These reflect the amount of edits we can see in examples in \cref{fig:syn_examples_motivation}. In \cref{sec:more_qual_examples}, we present additional examples of synthetic data from each method. Additionally, factoring in results from \cref{tab:ppp_ablate}, we find that larger edits corresponded to higher benefit of using PPP. In other words, the value of the pair label between a synthetic example and its real counterpart (indicating they are the same category) increases the more different the two are. 

\noindent\textbf{How well are classes preserved in generated synthetic data?} This is necessary, because PPP relies on the translations being label-preserving. To measure, we compute nearest class mean \textbf{(NCM) accuracy}. Specifically, we compute NCM accuracy in the CLIP feature space, over examples in the set $\D_{syn, B} = \{G_{A \rightarrow B} (\x) ; \x \in \D_A \}$ using the examples in $\D_B$ along with their ground truth category labels. In \cref{tab:domainnet_clip_scores} we report the mean of this and accuracy over examples in $\D_{syn, A}$ using real examples and category labels from $\D_A$. We find that Img2Img data has a high NCM accuracy which can possibly be explained by the lower amount of edit it makes to the real image during translation. While ELITE exhibits larger edits, it does not lag behind a lot in NCM accuracy, consequently making it a better synthetic data generator.

\noindent\textbf{How well does synthetic data mimic real data?} Ideally, the best synthetic data for performance is one that mimics the real data distribution (\ie it is the best replacement for the missing real data in \cref{fig:fig1}). Hence, we try to measure this using \textbf{Similarity to Real Target}. For an example $\x_{syn}^{B}$, with class label $y$ (note that this is the same as the class label of the real example used to generate $\x_{syn}^{B}$), we compute $f_{CLIP}(\x_{syn}^B)^\top \overline{f}_{CLIP}(\D_{B, y})$ (where $\overline{f}_{CLIP}(\D_{B, y})$ is the feature space mean of all examples from $\D_B$ with category label $y$), and find the average over all examples in $\D_{syn, B}$. In \cref{tab:domainnet_clip_scores} we report the mean of this and a similar average computed over examples in $\D_{syn, A}$ for each pair of domains. We find that ELITE generated data is closest to the real data, which is reflected in its efficacy as synthetic replacement for missing real data.

\vspace{-3mm}
\subsection{SynCDR performance with category overlap} \label{subsec:cat_overlap}

\begin{table}[]
\begin{center}

\caption{\textbf{Performance in the case of non-zero category overlap across domains in training data.} In the introduction, we motivated how synthetic data can come to the rescue in cases of missing similar category training examples across the two domains (\ie 0\% category overlap). In this experiment performed using the DomainNet dataset, we see that synthetic data (used to train SynCDR) is still useful to improve the performance of CDS~\cite{kim2021cds} even with non-zero category overlap.}

\scalebox{0.75}{
\begin{tabular}{@{}lrcccc@{}}
\toprule
\textbf{Method} & \multicolumn{1}{l}{\thead{\textbf{Category Overlap} \\ \textbf{in training data}}} & \textbf{C-P} & \textbf{C-S} & \textbf{P-S} & \textbf{Avg} \\
\midrule
\multirow{3}{*}{CDS} & 0\% & 28.6 ± 0.7 & 24.5 ± 0.6 & 30.5 ± 0.6 & 27.8 ± 0.6 \\
 & 50\% & 32.5 ± 0.6 & 29.3 ± 0.6 & 34.3 ± 0.9 & 32.1 ± 0.7 \\
 & 100\% & 37.1 ± 0.5 & 35.1 ± 0.4 & 40.5 ± 0.5 & 37.6 ± 0.5 \\
 \midrule
\multirow{3}{*}{In-domain ID} & 0\% & 30.7 ± 0.4 & 27.5 ± 0.2 & 31.8 ± 0.4 & 30.0 ± 0.4 \\
 & 50\% & 31.6 ± 0.3 & 28.3 ± 0.2 & 32.1 ± 0.2 & 30.6 ± 0.3 \\
 & 100\% & 31.6 ± 0.3 & 29.1 ± 0.1 & 32.8 ± 0.7 & 31.2 ± 0.5 \\
 \midrule
\multirow{3}{*}{\thead[l]{SynCDR \\ (w ELITE)}} & 0\% & 45.4 ± 0.7 & 44.2 ± 0.4 & 46.4 ± 1.3 & 45.3 ± 0.9 \\
 & 50\% & 48.9 ± 0.9 & 44.4 ± 0.4 & 45.7 ± 1.4 & 46.3 ± 1.0 \\
 & 100\% & 50.8 ± 0.4 & 48.5 ± 1.2 & 48.8 ± 1.1 & 49.4 ± 1.0 \\
 \bottomrule
\end{tabular}
}
\label{tab:diff_overlaps}
\vspace{-7mm}
\end{center}
\end{table}

In \cref{fig:fig1}, we motivated that synthetic data steps in to assist by replacing missing real training data in cases where similar categories of images are not available across domains. While this is the case when synthetic data is most effective, in this section, we show how much it can add to performance when training data across domains have overlapping categories.

In \cref{tab:diff_overlaps}, on each scenario of DomainNet we report performance of different methods with different amounts of category overlaps. The dataset has images from $126$ categories. $0\%$ overlap corresponds to results from \cref{tab:domainnet_results}, each domain having training data from $63$ categories. In the case of $50\%$ overlap, each of domains A and B has training data from 84 categories, 42 of which are shared by both domains. In the case of $100\%$ overlap, each domain has training data from each of the $126$ different categories. 

As discussed in the introduction, we can see the improved efficacy of CDS~\cite{kim2021cds} as it gets training examples from same categories across the two domains. This is in contrast to in-domain ID which improved much less. We can see that SynCDR still helps improve over CDS performance significantly in this case.

\vspace{-4mm}
\section{Conclusion} \label{sec:conclusion}
\vspace{-2mm}
Cross-domain retrieval models can be trained without labeled data but they rely heavily on presence of examples from same categories in both domains. In situations where this is unavailable, we show that they can perform poorly. In this paper, we presented a solution by replacing missing category examples with synthetic data. For this, we used label-preserving translation methods which could generate counterparts of real images in the opposite domain while preserving content, crucially, the semantic category. With such generated data, we get additional pair labels (indicating same category) for free, which we used during training by optimizing the pseudo positive pair PPP loss. We compared different synthetic data generators, including those that require specific training for a pair of domains as well as ones that can be prompted for translation and rely on pre-trained text-to-image generative diffusion models. Overall, with the best synthetic data, our SynCDR model could improve performance by $\sim$15\% on the DomainNet dataset and $\sim$6\% on the CUB dataset.

\noindent \textbf{Limitations.} The best synthetic data for SynCDR uses textual prompts which describe the domains. This may not work if domains involved cannot be described this way. Here, a method like Textual Inversion~\cite{gal2022image} could be used to learn the properties of a domain as a new token embedding. We evaluated this method on DomainNet in \cref{subsec:textinv} but found its performance to be poorer on average than using textual descriptions. Experimenting on tasks where textually describing domains is not possible, is a topic of future work.  

\noindent \textbf{Acknowledgments.} This work was funded by the National Science Foundation and the Hariri Institute at Boston University.

\bibliographystyle{splncs04}
\bibliography{main}

\break

\appendix
\section{Additional Results}

\subsection{Using Textual Inversion to learn domains} \label{subsec:textinv}
In the main paper under limitations, we mentioned that our synthetic data generation methods use simply a word or a phrase to represent the target domain (\cref{tab:prompts} lists these for different methods). This may not be possible in scenarios where this description is unavailable, \eg in data collected from specific sensors. We hypothesize that in such a case, Textual Inversion~\cite{gal2022image} can be used. It allows us to encode a concept exemplified via a set of input images in a new token by optimizing its embeddings such that a pre-trained and frozen Stable Diffusion model generates the input images conditioned on this new token. Note that this description obfuscates some details to provide an intuitive understanding. For a full and precise description of textual inversion, we refer readers to Sec 3 of \cite{gal2022image}. 

While in this paper we did not experiment with datasets where domains cannot be verbally described, we evaluated textually inverting domains on the DomainNet dataset. For textual inversion, we used a pre-trained Stable-Diffusion v1.4 model (which is the same version as used by ELITE), and encoded each domain (clipart, painting or sketch) using its unlabeled training images by using 5000 gradient steps, with a learning rate of $5 \times 10^{-4}$ and a batch size of 4. We used code available in \cite{von-platen-etal-2022-diffusers}. For generating a synthetic image in domain A, we use the prompt ``An image of S in the style of $<d_A>$'', where S is the token corresponding to the input image object encoded by ELITE and $<d_A>$ corresponds to the textually inverted token encoding the properties of domain A.

\begin{figure*}[]
    \centering
    \includegraphics[width=0.95\linewidth,trim=0cm 0cm 0cm 0cm,clip]{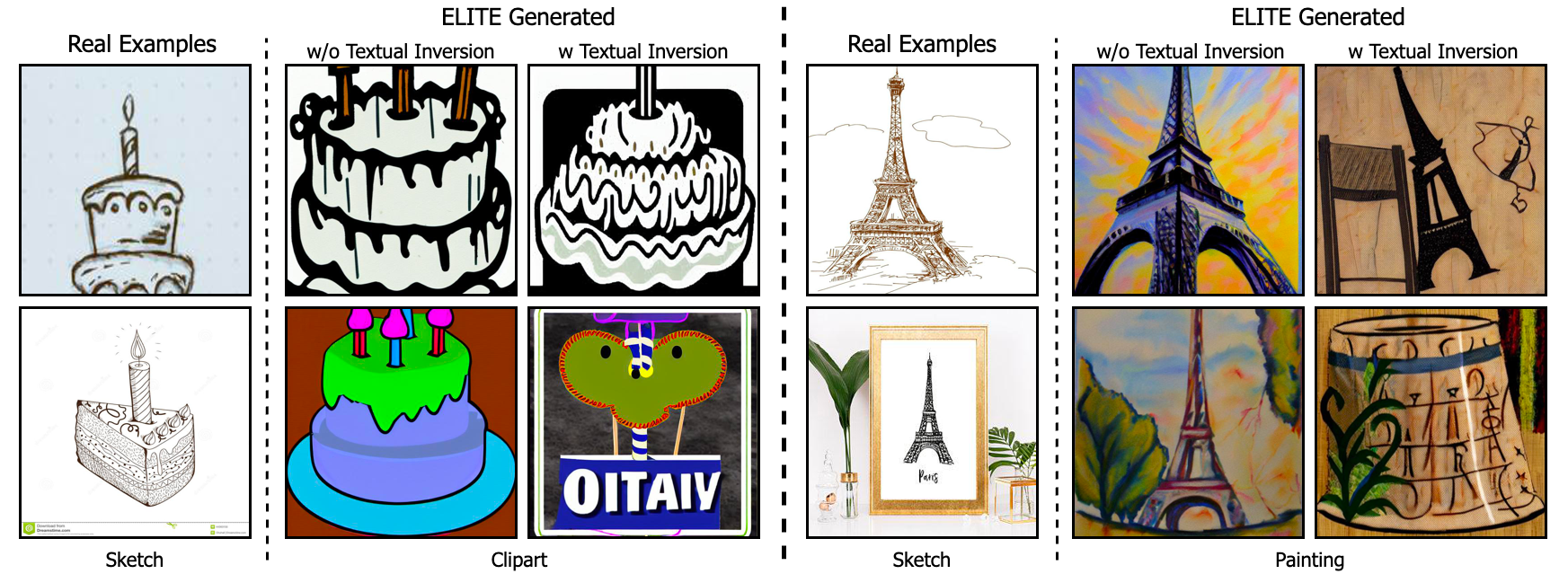}
    \caption{{\small \textbf{ELITE Generated examples with and without textual inversion.} When generating paintings or cliparts from sketches we found that using a textual inversion token encoding the domain properties leads to poorer category retention in the generated image, and hence leading to poorer performance in general. Textual Inversion can however be useful in scenarios when the domain cannot be textually described.}}
    \label{fig:textinv_examples} \vspace{-1mm}
\end{figure*}

\begin{table}[]
\begin{center}
\caption{\textbf{Using Textual Inversion to learn Domain Properties.} We see that when we use a textually inverted token (see \cref{subsec:textinv} and \cite{gal2022image}) that encodes domain properties for generating synthetic data with ELITE, cross-domain retrieval performance of SynCDR is similar or decreases. This happens possibly because Stable Diffusion's understanding of common domains such as paintings, clipart and sketches is quite good. Additionally, as discussed in \cref{subsec:textinv}, we see that in cases of generating paintings or clipart from sketches, using textual inversion is less accurate in preserving category of the input example. See examples in \cref{fig:textinv_examples}}
\scalebox{0.85}{
\begin{tabular}{@{}lcccc@{}}
\toprule
\textbf{Method} & \textbf{C-P} & \textbf{C-S} & \textbf{P-S} & \textbf{Avg} \\
\midrule
SynCDR (ELITE) & 45.4 ± 0.7 & 44.2 ± 0.4 & 46.4 ± 1.3 & 45.3 ± 0.9 \\
\thead[l]{SynCDR (ELITE \\ with textual inversion)} & 45.1 ± 0.5 & 41.8 ± 0.6 & 43.0 ± 0.8 & 43.3 ± 0.7 \\
\bottomrule
\end{tabular}
}

\label{tab:textinv_results}
\end{center}
\end{table}

From the results in \cref{tab:textinv_results}, we see that textually inverting a domain's properties leads to similar or poorer results overall. It is poorer in the case when sketch is involved as a domain. On digging deeper, we find that when generating paintings or cliparts from sketches, using textual inversion is poorer at preserving the category of the input sketch example. For instance, NCM accuracy (as defined in Sec. 4.3 of the main paper) of synthetic paintings generated with textual inversion is 0.54 compared to 0.61 without (see some examples of such cases in \cref{fig:textinv_examples}). This suggests the generation process somehow get skewed more to pay attention on only the style in such cases. We leave further investigation of this method to future work.

\subsection{Distilling large CLIP models for cross-domain retrieval} \label{subsec:distill}

As mentioned in Sec. 4.1 (main paper) and as shown in \cref{tab:clip_distill}, image embeddings from large CLIP transformer models are robust across domains, but are expensive to obtain on smaller memory devices at inference time. In the top section of \cref{tab:clip_distill}, we report the performances of two of these CLIP pre-trained models. The CLIP RN-50 is a larger version of the ResNet-50 model with $\sim$40M parameters compared to the latter's $\sim$25M. The CLIP ViT-L is a large transformer model, which provides highly robust image embeddings at test time, but is expensive to store and run with $\sim$305M parameters.

We additionally note that ELITE's image encoder, which generates the token embedding encoding an input image concept, uses the CLIP ViT-L model. ELITE can hence be interpreted as a way of distilling the information in the CLIP ViT-L via synthetic training data. An alternative approach would be to distill the CLIP ViT-L model based on its feature similarities between pairs of images in the training data. This approach is described next. 

CLIP-distill (\cref{tab:clip_distill}) distills a CLIP ViT-L into a ResNet-50. Its optimization criterion is similar to Sun~\etal~\cite{sun2023dime}, but uses only unlabeled images available to us as training data. More precisely, during training we are given a teacher (CLIP ViT-L) feature extractor $g : \mc{X} \rightarrow \bS^{d}$ (recall $\bS^d$ is the $d$-dimensional unit hypersphere) and a student model $f: \mc{X} \rightarrow \bS^{d}$. Additionally for $m$-sized minibatches of images $X_A$ from the domain A and $X_B$ from domain B, let us define $p_f(\x, X) = \text{softmax}(f(\x)^\top f(X))$ (where $\x \in \mc{X}$ and we overloaded notation for the batch of $m$ images $X$ so that $f(X) = [f(\x) ; \x \in X] \in \R^{d \times m}$ )

\begin{align} \label{eq:l_distill}
    L_{distill} &(X_A, X_B) \nonumber \\ 
    &= \frac{1}{m} \sum_{\x^{A} \in X_A} D_{KL}(p_f(\x^{A}, X_B) || p_g(\x^{A}, X_B)) \nonumber \\
    &+ \frac{1}{m} \sum_{\x^{A} \in X_A} D_{KL}(p_f(\x^{A}, X_A) || p_g(\x^{A}, X_A)) \nonumber \\
    &+ \frac{1}{m} \sum_{\x^{B} \in X_B} D_{KL}(p_f(\x^{B}, X_A) || p_g(\x^{B}, X_A)) \nonumber \\
    &+ \frac{1}{m} \sum_{\x^{B} \in X_B} D_{KL}(p_f(\x^{B}, X_B) || p_g(\x^{B}, X_B))
\end{align}

where $D_{KL}$ is the KL divergence between two distributions. In essence, we attempt to distill the similarity predictions made by the teacher CLIP model $g$ for training images across domains (terms 1 and 3 of \cref{eq:l_distill}), and those within the same domain (terms 2 and 4 of \cref{eq:l_distill}).

From the results, we see that distilling the large CLIP model helps with performance compared to ID or CDS, but SynCDR (\ie training with synthetic data) is a much more effective approach in terms of performance.

\begin{table}[]
\begin{center}
\caption{\textbf{CLIP Distillation over different scenarios of DomainNet.} The top two rows contain the performance of using pre-trained CLIP models at inference time (CLIP ViT-L is a large transformer with $\sim$305M parameters, and CLIP RN-50 is a resnet-50 variant with $\sim$40M parameters). The next two rows, CDS and ID use train a ResNet-50 backbone in a self-supervised manner on the unlabeled training images available in the two domains. CLIP-distill distills the similarity predictions from a CLIP ViT-L model into a ResNet-50 using the same data. ELITE uses a CLIP ViT-L backbone in its image encoder for generating synthetic examples. As seen from the results, using these for training leads to a better performing model than distilling via feature similarities.}

\scalebox{0.85}{
\begin{tabular}{@{}lcccc@{}}
\toprule
\textbf{Model} & \textbf{C-P} & \textbf{C-S} & \textbf{P-S} & \textbf{Avg} \\
\midrule
CLIP (pt) - RN-50 & 35.3 & 39.9 & 39.5 & 38.2 \\
CLIP (pt) - ViT-L & 66.3 & 74.8 & 70.1 & 70.4 \\
\midrule
CDS & 28.6 ± 0.7 & 24.5 ± 0.6 & 30.5 ± 0.6 & 27.8 ± 0.6 \\
In-domain ID & 30.7 ± 0.4 & 27.5 ± 0.2 & 31.8 ± 0.4 & 30.0 ± 0.4 \\
\midrule
CLIP-distill & 33.2 ± 0.5 & 32.3 ± 0.5 & 33.4 ± 0.4 & 33.0 ± 0.5 \\
\midrule
SynCDR (with ELITE) & 45.4 ± 0.7 & 44.2 ± 0.4 & 46.4 ± 1.3 & 45.3 ± 0.9 \\
\bottomrule
\end{tabular}
}

\label{tab:clip_distill}
\end{center}
\end{table}

\subsection{Full Retrieval Results} \label{subsec:full_ret_results}

\begin{table*}[]
\centering
\caption{\textbf{Cross Domain Retrieval performance (Prec@1, 5, 15) across different scenarios of DomainNet.} Here we additionally report precisions at different number of retrieved examples for all models from Tab. 1 (main paper).}
\begin{subtable}{\textwidth}
\centering
\scalebox{0.55}{
\begin{tabular}{@{}ll|ccc|ccc|ccc@{}}
\toprule
\multirow{2}{*}{\textbf{Model}} & \multirow{2}{*}{\textbf{Synthetic Data}} & \multicolumn{3}{c|}{\textbf{Clipart   - Painting}} & \multicolumn{3}{c|}{\textbf{Clipart   - Sketch}} & \multicolumn{3}{c}{\textbf{Painting   - Sketch}}\\
 &  & \textbf{Prec@1} & \textbf{Prec@5} & \textbf{Prec@15} & \textbf{Prec@1} & \textbf{Prec@5} & \textbf{Prec@15} & \textbf{Prec@1} & \textbf{Prec@5} & \textbf{Prec@15} \\
\midrule
ImageNet (pt) & \multicolumn{1}{c|}{-} & 27.4 & 23.4 & 19.2 & 24.6 & 20.4 & 16.8 & 29.2 & 25.5 & 21.5 \\
CDS & \multicolumn{1}{c|}{-} & 28.6 ± 0.7 & 24.5 ± 0.7 & 20.5 ± 0.6 & 24.5 ± 0.6 & 20.8 ± 0.6 & 17.6 ± 0.6 & 30.5 ± 0.6 & 26.9 ± 0.6 & 23.0 ± 0.6 \\
ID & \multicolumn{1}{c|}{-} & 30.7 ± 0.4 & 26.5 ± 0.4 & 22.2 ± 0.3 & 27.5 ± 0.2 & 23.3 ± 0.2 & 19.3 ± 0.3 & 31.8 ± 0.4 & 28.1 ± 0.3 & 24.0 ± 0.3 \\
\multirow{4}{*}{SynCDR (Ours)} & CUT & 38.8 ± 0.8 & 34.2 ± 0.7 & 30.0 ± 0.7 & 40.1 ± 0.4 & 36.5 ± 0.4 & 32.7 ± 0.4 & 44.7 ± 0.6 & 41.2 ± 0.5 & 37.2 ± 0.5 \\
 & Img2Img & 38.2 ± 0.7 & 33.3 ± 0.6 & 28.7 ± 0.4 & 34.5 ± 0.7 & 30.8 ± 0.6 & 27.2 ± 0.5 & 37.4 ± 0.5 & 33.9 ± 0.5 & 30.1 ± 0.4 \\
 & InstructPix2Pix & 41.6 ± 0.6 & 36.6 ± 0.6 & 32.0 ± 0.5 & 39.8 ± 0.7 & 35.9 ± 0.7 & 32.1 ± 0.6 & 42.5 ± 0.7 & 38.8 ± 0.6 & 34.6 ± 0.5  \\
 & ELITE & \textbf{45.4 ± 0.7} & \textbf{41.6 ± 0.6} & \textbf{37.1 ± 0.6} & \textbf{44.2 ± 0.4} & \textbf{41.1 ± 0.3} & \textbf{37.6 ± 0.2} & \textbf{46.4 ± 1.3} & \textbf{43.9 ± 1.3} & \textbf{40.3 ± 1.2} \\
\bottomrule
\vspace{-2mm}
\end{tabular}
}
\end{subtable}

\begin{subtable}{\textwidth}
\centering
\scalebox{0.58}{
    \begin{tabular}{@{}ll|ccc@{}}
    \toprule
    \multirow{2}{*}{\textbf{Model}} & \multirow{2}{*}{\textbf{Synthetic Data}} & \multicolumn{3}{c}{\textbf{Average}} \\
     &  & \textbf{Prec@1} & \textbf{Prec@5} & \textbf{Prec@15} \\
    \midrule
    ImageNet (pt) & \multicolumn{1}{c|}{-} & 27.1 & 23.1 & 19.2 \\
    CDS & \multicolumn{1}{c|}{-} & 27.8 ± 0.6 & 24.1 ± 0.6 & 20.4 ± 0.6 \\
    ID & \multicolumn{1}{c|}{-} & 30.0 ± 0.4 & 26.0 ± 0.3 & 21.8 ± 0.3 \\
    \multirow{4}{*}{SynCDR (Ours)} & CUT & 41.2 ± 0.6 & 37.3 ± 0.6 & 33.3 ± 0.5 \\
     & Img2Img & 36.7 ± 0.6 & 32.7 ± 0.6 & 28.7 ± 0.5 \\
     & InstructPix2Pix & 41.3 ± 0.7 & 37.1 ± 0.6 & 32.9 ± 0.5 \\
     & ELITE & \textbf{45.3 ± 0.9} & \textbf{42.2 ± 0.9} & \textbf{38.3 ± 0.8} \\
     \bottomrule
    \end{tabular}
}
\end{subtable}
\label{tab:domainnet_all_precs}
\end{table*}

\begin{table}[]
\begin{center}
\caption{\textbf{Cross Domain Retrieval performance (Prec@1, 5, 15) for CUB$\leftrightarrow$CUB-Paintings.} Here we additionally report precisions at different number of retrieved examples for all models from Tab. 2 (main paper).}
\scalebox{0.8}{
\begin{tabular}{@{}ll|ccc@{}}
\toprule
\multirow{2}{*}{\textbf{Model}} & \multirow{2}{*}{\textbf{Synthetic Data}} & \multicolumn{3}{c}{\textbf{Painting   - Real}} \\
 &  & \textbf{Prec@1} & \textbf{Prec@5} & \textbf{Prec@15} \\
\midrule
ImageNet (pt) & \multicolumn{1}{c|}{-} & 20.5 & 17.0 & 14.7 \\
CDS & \multicolumn{1}{c|}{-} & 22.0 ± 0.8 & 18.2 ± 0.7 & 16.1 ± 0.6 \\
ID & \multicolumn{1}{c|}{-} & 21.4 ± 1.2 & 18.3 ± 1.0 & 16.0 ± 0.6 \\
\multirow{4}{*}{SynCDR (Ours)} & CUT & 23.2 ± 0.5 & 19.3 ± 0.4 & 16.7 ± 0.4 \\
 & Img2Img & 22.0 ± 0.5 & 18.2 ± 0.6 & 16.0 ± 0.5 \\
 & InstructPix2Pix & 21.5 ± 0.7 & 17.8 ± 0.3 & 15.7 ± 0.3 \\
 & ELITE & \textbf{28.2 ± 0.4} & \textbf{23.8 ± 0.2} & \textbf{21.0 ± 0.2} \\
 \bottomrule
\end{tabular}
} %
\label{tab:cub_all_precs}
\end{center}
\end{table}

\begin{table*}[]
\centering
\caption{\textbf{Cross Domain Retrieval performance (Prec@1, 5, 15) in different scenarios of Office-Home.} Here we additionally report precisions at different number of retrieved examples for all models from Tab 3 (main paper).}
\label{tab:office_home_all_precs}
\begin{subtable}{\textwidth}
\centering
\scalebox{0.55}{
\begin{tabular}{@{}ll|ccc|ccc|ccc@{}}

\toprule
\multirow{2}{*}{\textbf{Model}} & \multirow{2}{*}{\textbf{Synthetic Data}} & \multicolumn{3}{c|}{\textbf{Art -   Clipart}} & \multicolumn{3}{c|}{\textbf{Art -   Product}} & \multicolumn{3}{c}{\textbf{Art -   Real}} \\ 
 &  & \textbf{Prec@1} & \textbf{Prec@5} & \textbf{Prec@15} & \textbf{Prec@1} & \textbf{Prec@5} & \textbf{Prec@15} & \textbf{Prec@1} & \textbf{Prec@5} & \textbf{Prec@15} \\ \midrule
ImageNet (pt) & \multicolumn{1}{c|}{-} & 34.5 & 29.3 & 24.4 & 48.6 & 40.7 & 34.5 & 54.6 & 47.8 & 42.8 \\
CDS & \multicolumn{1}{c|}{-} & 36.5 ± 0.3 & 31.4 ± 0.2 & 26.7 ± 0.2 & 50.0 ± 0.4 & 42.1 ± 0.1 & 36.3 ± 0.2 & 54.9 ± 0.3 & 48.9 ± 0.1 & 43.6 ± 0.1 \\
In-domain ID & \multicolumn{1}{c|}{-} & 36.3 ± 0.3 & 31.2 ± 0.2 & 26.6 ± 0.1 & 50.0 ± 0.4 & 41.9 ± 0.1 & 36.1 ± 0.1 & 54.7 ± 0.3 & 48.9 ± 0.1 & 43.5 ± 0.1 \\
\multirow{4}{*}{SynCDR (Ours)} & CUT & 38.3 ± 0.4 & \textbf{33.3 ± 0.2} & 28.7 ± 0.1 & 50.8 ± 0.2 & \textbf{42.6 ± 0.1} & 37.0 ± 0.1 & \textbf{55.5 ± 0.6} & \textbf{49.7 ± 0.2} & 44.7 ± 0.1 \\
 & Img2Img & 36.4 ± 0.3 & 31.4 ± 0.1 & 26.8 ± 0.1 & 49.9 ± 0.3 & 42.2 ± 0.2 & 36.4 ± 0.1 & 54.9 ± 0.5 & 49.2 ± 0.2 & 44.0 ± 0.2 \\
 & InstructPix2Pix & \textbf{38.6 ± 0.5} & 33.0 ± 0.2 & 28.6 ± 0.3 & 50.2 ± 0.2 & 42.3 ± 0.2 & 36.6 ± 0.2 & 54.7 ± 0.3 & 49.3 ± 0.1 & 43.9 ± 0.1 \\
 & ELITE & 38.3 ± 0.5 & 33.1 ± 0.2 & \textbf{29.1 ± 0.2} & \textbf{51.3 ± 0.3} & 42.5 ± 0.1 & \textbf{37.2 ± 0.1} & 55.4 ± 0.3 & \textbf{49.7 ± 0.2} & \textbf{44.8 ± 0.1} \\ 
 \bottomrule
 \toprule
\multirow{2}{*}{\textbf{Model}} & \multirow{2}{*}{\textbf{Synthetic Data}} & \multicolumn{3}{c|}{\textbf{Clipart - Product}} & \multicolumn{3}{c|}{\textbf{Clipart - Real}} & \multicolumn{3}{c}{\textbf{Product - Real}} \\
 &  & \textbf{Prec@1} & \textbf{Prec@5} & \textbf{Prec@15} & \textbf{Prec@1} & \textbf{Prec@5} & \textbf{Prec@15} & \textbf{Prec@1} & \textbf{Prec@5} & \textbf{Prec@15} \\ \midrule
ImageNet (pt) & \multicolumn{1}{c|}{-} & 42.2 & 35.9 & 29.4 & 43.9 & 38.0 & 30.2 & 68.6 & 60.7 & 49.9 \\
CDS & \multicolumn{1}{c|}{-} & 43.0 ± 0.2 & 38.0 ± 0.2 & 31.2 ± 0.1 & 45.2 ± 0.3 & 40.6 ± 0.2 & 32.5 ± 0.2 & 69.1 ± 0.2 & 62.5 ± 0.2 & 52.2 ± 0.2 \\
In-domain ID & \multicolumn{1}{c|}{-} & 43.4 ± 0.4 & 37.9 ± 0.2 & 31.0 ± 0.1 & 45.4 ± 0.4 & 40.4 ± 0.2 & 32.4 ± 0.2 & 69.0 ± 0.3 & 62.4 ± 0.1 & 52.1 ± 0.1 \\
\multirow{4}{*}{SynCDR (Ours)} & CUT & 43.3 ± 0.5 & 38.0 ± 0.4 & 31.3 ± 0.3 & 46.6 ± 0.3 & 41.8 ± 0.3 & \textbf{34.5 ± 0.2} & \textbf{69.3 ± 0.3} & 62.8 ± 0.1 & 53.0 ± 0.1 \\
 & Img2Img & 43.4 ± 0.4 & 38.1 ± 0.2 & 31.2 ± 0.2 & 45.2 ± 0.4 & 40.4 ± 0.3 & 32.6 ± 0.3 & 69.0 ± 0.3 & 62.6 ± 0.1 & 52.5 ± 0.1 \\
 & InstructPix2Pix & 43.9 ± 0.2 & 39.2 ± 0.2 & 32.6 ± 0.2 & \textbf{46.7 ± 0.4} & \textbf{42.0 ± 0.2} & 34.0 ± 0.3 & 69.0 ± 0.2 & 62.5 ± 0.1 & 52.4 ± 0.1 \\
 & ELITE & \textbf{44.1 ± 0.6} & \textbf{39.3 ± 0.4} & \textbf{32.7 ± 0.4} & 46.3 ± 0.5 & 41.6 ± 0.3 & 33.8 ± 0.5 & \textbf{69.3 ± 0.3} & \textbf{62.9 ± 0.3} & \textbf{53.1 ± 0.4} \\ \bottomrule

 \end{tabular}
 }
 \end{subtable}
 \begin{subtable}{\textwidth}
 \centering
 \vspace{0.1mm}
 \scalebox{0.58}{
    \begin{tabular}{@{}ll|ccc@{}}
    \toprule
    \multirow{2}{*}{\textbf{Model}} & \multirow{2}{*}{\textbf{Synthetic Data}} & \multicolumn{3}{c}{\textbf{Average}} \\ 
     &  & \textbf{Prec@1} & \textbf{Prec@5} & \textbf{Prec@15} \\ \midrule
    ImageNet (pt) & \multicolumn{1}{c|}{-} & 48.7 & 42.1 & 35.2 \\
    CDS & \multicolumn{1}{c|}{-} & 49.8 ± 0.3 & 43.9 ± 0.2 & 37.1 ± 0.2 \\
    In-domainID & \multicolumn{1}{c|}{-} & 49.8 ± 0.3 & 43.8 ± 0.2 & 36.9 ± 0.1 \\
    \multirow{4}{*}{SynCDR (Ours)} & CUT & 50.6 ± 0.4 & 44.7 ± 0.2 & 38.2 ± 0.2 \\
     & Img2Img & 49.8 ± 0.4 & 44.0 ± 0.2 & 37.3 ± 0.2 \\
     & InstructPix2Pix & 50.5 ± 0.3 & 44.7 ± 0.2 & 38.0 ± 0.2 \\
     & ELITE & \textbf{50.8 ± 0.4} & \textbf{44.8 ± 0.3} & \textbf{38.4 ± 0.3}  \\
     \bottomrule
    \end{tabular}
}
\end{subtable}

\end{table*}

In \cref{tab:domainnet_all_precs,,tab:cub_all_precs,,tab:office_home_all_precs}, we report Precs @1, 5 and 15 retrieved images as done in \cite{kim2021cds}. We observe largely similar trends using Prec@5 and Prec@15 as those using Prec@1, and results are reported for completeness.

\subsection{Feature Visualization} \label{subsec:feat_viz}

\begin{figure}[]
    \centering
    \includegraphics[width=0.8\linewidth,trim=0cm 0cm 0cm 0cm,clip]{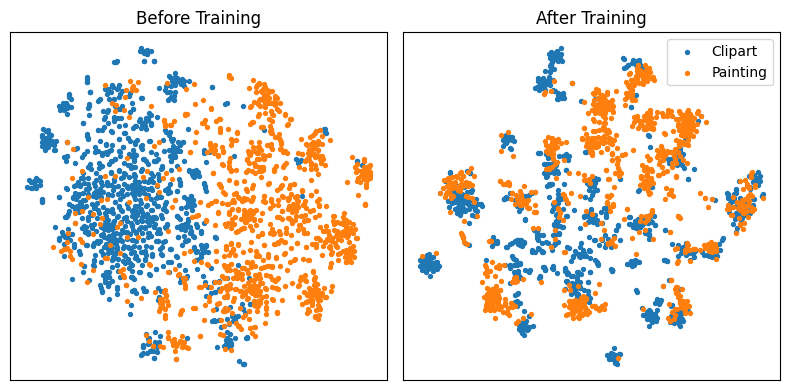}
    \caption{{\small Feature visualization before and after training. The examples (from test set of DomainNet Clipart and Painting) get better clustered and more aligned across domains after training.}}
    \label{fig:domainnet_c2p_tsne} 
\end{figure}

\cref{fig:domainnet_c2p_tsne} shows t-SNE visualization of the output features from the network before and after SynCDR training. The two domains in the plot are Clipart and Painting from the DomainNet dataset and we used 1000 points from the test set of  each domain picked at random for the visualization. We see that features get better clustered class-wise and are more aligned across domains after SynCDR training.

\section{Synthetic Data Generation Details} \label{sec:syn_details}

\subsection{Contrastive Unpaired Translation (CUT)}

CUT~\cite{park2020contrastive} trains a generator that can translate an input image to visually resemble a different domain. It can be trained using images from the two visual domains without the need for paired illustrations, making it viable for our use-case. CUT relies on optimizing a combination of a GAN loss and patchwise contrastive losses where the latter reuses features from the encoder of the image generator (which is a composition of a convolutional encoder and decoder) and uses negative patches from within the input image, resulting in a simple efficient approach without many components. 

For our experiments, we used two CUT generators from code provided by Park~\etal~\cite{park2020contrastive} trained for translation in either direction on the unlabeled training images in each of the two domains. This is the relatively slower higher quality version of CUT compared to FastCUT, a faster variant that Park~\etal~\cite{park2020contrastive} also provide. For training each generator, we used a batch size of 4, and trained for 400 epochs (where each epoch was clipped to a maximum of 250 steps). While after training, image generation is fast (90 ms compared to few seconds for diffusion sampling methods---as described next), the training itself is relatively slow, taking approximately 20 hrs on an NVIDIA Tesla V100 gpu.

\subsection{Img2Img/SDEdit}

\begin{figure*}[t]
    \centering
    \includegraphics[width=0.87\linewidth,trim=0cm 0cm 0cm 0cm,clip]{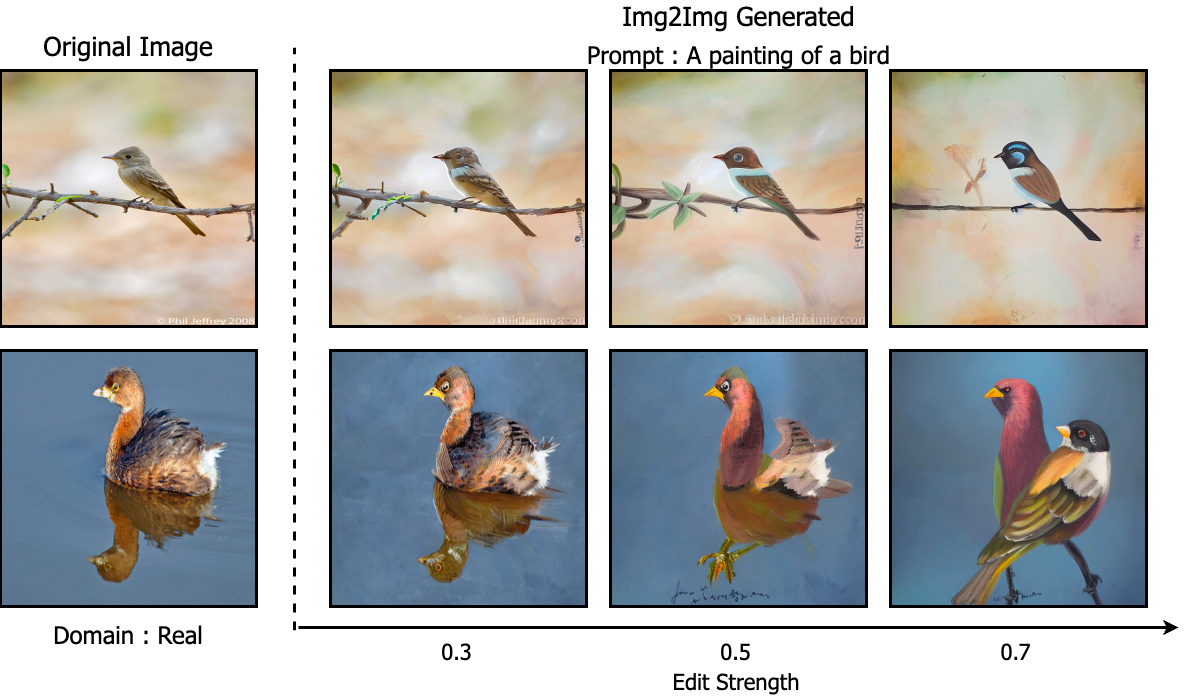}
    \caption{{\small \textbf{Different Edit Strengths for translation using Img2Img in the CUB dataset.} We see that higher edit strengths can make the output resemble paintings more, but they can drastically change the image contents such that the category is not preserved.}}
    \label{fig:diff_strengths_img2img} \vspace{-1mm}
\end{figure*}

SDEdit~\cite{meng2021sdedit} was a method first proposed for editing images before the advent of open large scale pre-trained text-to-image diffusion models. With a pre-trained denoising diffusion model the approach is straightforward, following the diffusion noising process on an input image, and then subsequently denoising it. What generates the edit is following the noising process partially based on an \textsl{edit strength} $\in [0, 1]$, such that not all information in the input image is lost. At the extremes, the value 0 for this parameter leads to no edit, and the value 1 leads to losing all input image information.

With a text-guided generative model like Stable Diffusion~\cite{rombach2022high}, this approach readily facilitates text-guided image editing using prompts. For our experiments, we use the implementation of \texttt{StableDiffusionImg2ImgPipeline} in \cite{von-platen-etal-2022-diffusers} with the pre-trained model \texttt{runwayml/stable-diffusion-v1-5}. We used 50 diffusion steps, a guidance scale of 10 (these values being somewhat standard for Stable Diffusion image generation) and chose the final edit strength as 0.3, after computing validation accuracy for SynCDR using data for 3 different values : 0.3, 0.5 and 0.7. The prompts we used for translating to each domain are mentioned in \cref{tab:prompts}. Since edit strength determines number of diffusion steps, the speed of the process depends on the edit strength parameter. For 0.3, we generated synthetic data at $\sim$2.8 seconds per image using an Nvidia Tesla V100 GPU, resulting in a total generation time of 6.7 hrs for our largest training dataset (painting domain of domainnet with $\sim$8600 images).

In \cref{fig:diff_strengths_img2img}, we show examples with different edit strength values. We notice in that example, that a higher strength can make a generated image look more like a painting, but at the cost of typically not retaining the input image category.

\subsection{InstructPix2Pix}

\begin{figure*}[t]
    \centering
    \includegraphics[width=0.87\linewidth,trim=0cm 0cm 0cm 0cm,clip]{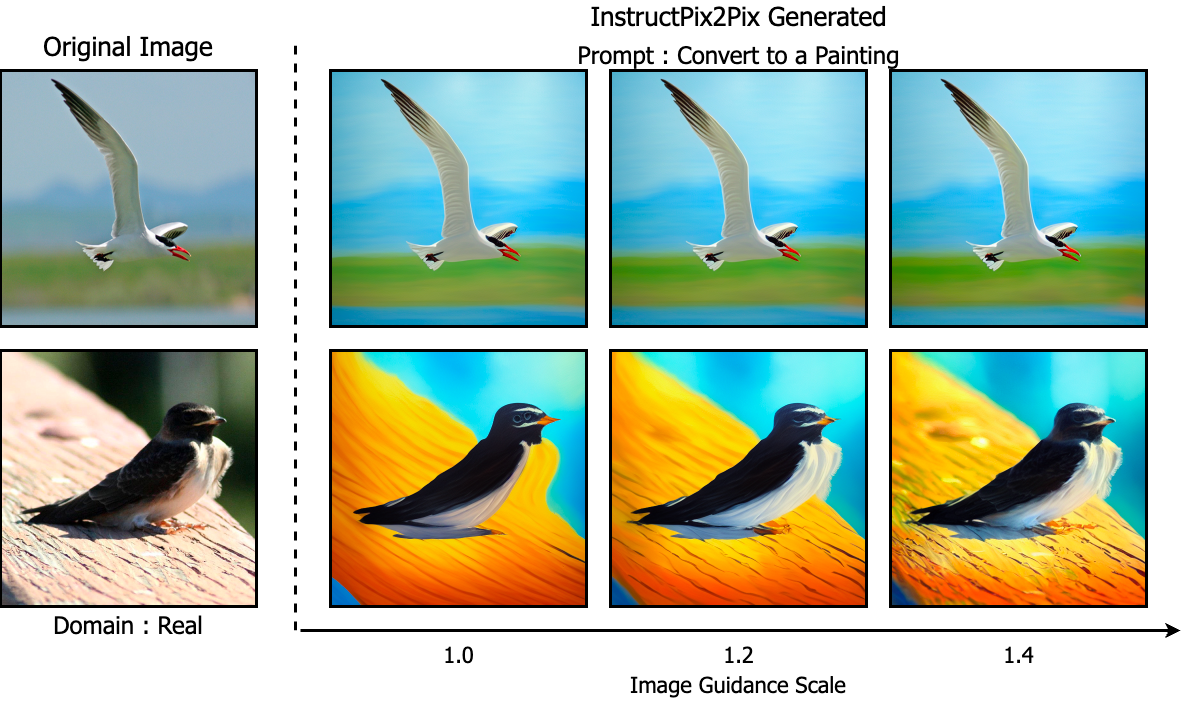}
    \caption{{\small \textbf{Different image guidance scales for edits made using InstructPix2Pix in the CUB dataset.}} Compared to Img2Img, this is better across multiple guidance scales at maintaining the input image category. However, at low guidance scales, it can still make errors in this such as that shown in the second row.}
    \label{fig:diff_strengths_ip2p} \vspace{-1mm}
\end{figure*}

InstructPix2Pix~\cite{brooks2023instructpix2pix} is a denoising diffusion model that can be conditioned on both an image and a text-prompt trained to carry out an edit to the input image, specified by the prompt. To control the amount of edit, we can use a parameter $image\_guidance\_scale \ge 1$, larger values of which lead to the output image looking more like the input image.

For our experiments, we used \texttt{StableDiffusionInstructPix2PixPipeline} from \cite{von-platen-etal-2022-diffusers}, with the pre-trained model provided by \cite{brooks2023instructpix2pix} in \break \texttt{timbrooks/instruct-pix2pix}. We used 50 diffusion steps and a guidance scale of 10, and for setting the $image\_guidance\_scale$ we used validation accuracy across 4 different values of generation : [1, 1.2, 1.4, 1.6], finally choosing 1.2. The text prompts used to provide the edit instructions for each domain are in \cref{tab:prompts}. Generation speed for this method is relatively slower than Img2Img at around 9 seconds per image resulting in a total generation time of 21 hrs on an NVIDIA Tesla V100 GPU for our largest training dataset (painting domain of domainnet with $\sim$8600 images). 

\cref{fig:diff_strengths_ip2p} shows examples of InstructPix2Pix edits made with different values of $image\_guidance\_scale$ to convert a realistic photo of a bird to a painting. We notice that InstructPix2Pix does a decent job of editing across the different scales, however on occasion a smaller guidance can lead to large edits which may not preserve the bird category (for \eg $image\_guidance\_scale = 1$ in the second row).

\subsection{ELITE}

ELITE~\cite{wei2023elite} is a method for fast personalized image generation from Stable Diffusion. Its goal is to learn an object from a single image and generate it in different contexts and styles based on a text prompt. This, ELITE does by encoding an input image into a token embedding that encodes the object into the dictionary of Stable Diffusion's text encoder. Here we note that both ELITE and Textual Inversion~\cite{gal2022image} are personalized generation methods which can learn concepts from one or a few input images in the form of a new token embedding. While textual inversion does this via optimization for each new object/concept, which is slower, ELITE can do the same by using a trained encoder module which generates a new concept token embedding in one forward pass. 

The full ELITE method uses both a global and a local image encoding module, where the latter is useful for better preservation of details from the input object (interested readers may see Figure 6 of \cite{wei2023elite}). In our experiments, we utilize only the global module since the latter additionally requires a segmentation mask around the primary object in the input image. For generation, we used the implementation of \cite{wei2023elite}, which uses an image encoder based on a CLIP pre-trained ViT-L model from \cite{radford2021learning}, and Stable Diffusion v1.4 for generation. The different prompts we used for generation are listed in \cref{tab:prompts}. In these prompts, ``S'' corresponds to the special token that is output by ELITE's image encoder representing the input object. Generation speed was about the same as InstructPix2Pix at around 9s per image on an NVIDIA Tesla V100 gpu resulting in 21 hrs total generation time for domainnet paintings domain (with $\sim$8600 images).

\begin{table*}[]
\begin{center}
\caption{\textbf{Prompts used for synthetic data generation with different approaches.} In the case of ELITE, ``S'' refers to the special token representing the input object, whose embedding is output by ELITE's image encoder.}
\label{tab:prompts}
\scalebox{0.55}{
\begin{tabular}{@{}lllll@{}}
\toprule
\textbf{Dataset} & \textbf{Domain} & \multicolumn{1}{c}{\textbf{Img2Img Prompt}} & \multicolumn{1}{c}{\textbf{InstructPix2Pix   Prompt}} & \multicolumn{1}{c}{\textbf{ELITE Prompt}} \\
\midrule
\multirow{3}{*}{DomainNet} & Clipart & A clipart image of an object. & Convert to a clipart image. & A clipart image of S \\
 & Painting & A painting of an object. & Convert to a painting. & A painting of S \\
 & Sketch & A pencil/charcoal sketch of an object. & Convert to a pencil/charcoal sketch. & A pencil/charcoal sketch of S \\
 \midrule
\multirow{2}{*}{CUB} & Painting & A painting of a bird. & Convert to a painting. & A painting of S \\
 & Real & A realistic photo of a bird. & Convert to a realistic photo. & A realistic photo of S \\
 \midrule
\multirow{4}{*}{Office-Home} & Art & A painting of an object. & Convert to a painting. & A painting of S \\
 & Clipart & A clipart image of an object. & Convert to a clipart image. & A clipart image of S \\
 & Product & A photo of an object without a background. & Convert to a photo without a background. & An image of S without a background \\
 & Real & A realistic photo of an object. & Convert to a realistic photo. & A realistic photo of S \\
 \bottomrule
\end{tabular}
}
\end{center}
\end{table*}

\section{Other Training Details} \label{sec:train_details}
For training SynCDR, we use most of the same hyperparameters as CDS~\cite{kim2021cds}. Our model architecture is a ResNet-50 backbone followed by a fully connected layer resulting in an output feature dimension of 512. Before training, the backbone is initialized to Imagenet pre-trained weights. We use SGD with a learning rate of 0.003, with a batch size of 32 and momentum 0.9 for training SynCDR using a combination of the CDS and PPP losses as described in Sec. 3 of the main paper. We trained all models for 15 epochs validating after each epoch and do early stopping based on validation set Prec@1.

\section{Code}
The code for running SynCDR has been attached in a zip folder with the supplement. We intend to make it publicly available on github with the final version of the paper.

\begin{figure*}[]
    \centering
    \includegraphics[width=0.95\linewidth,trim=0cm 0cm 0cm 0cm,clip]{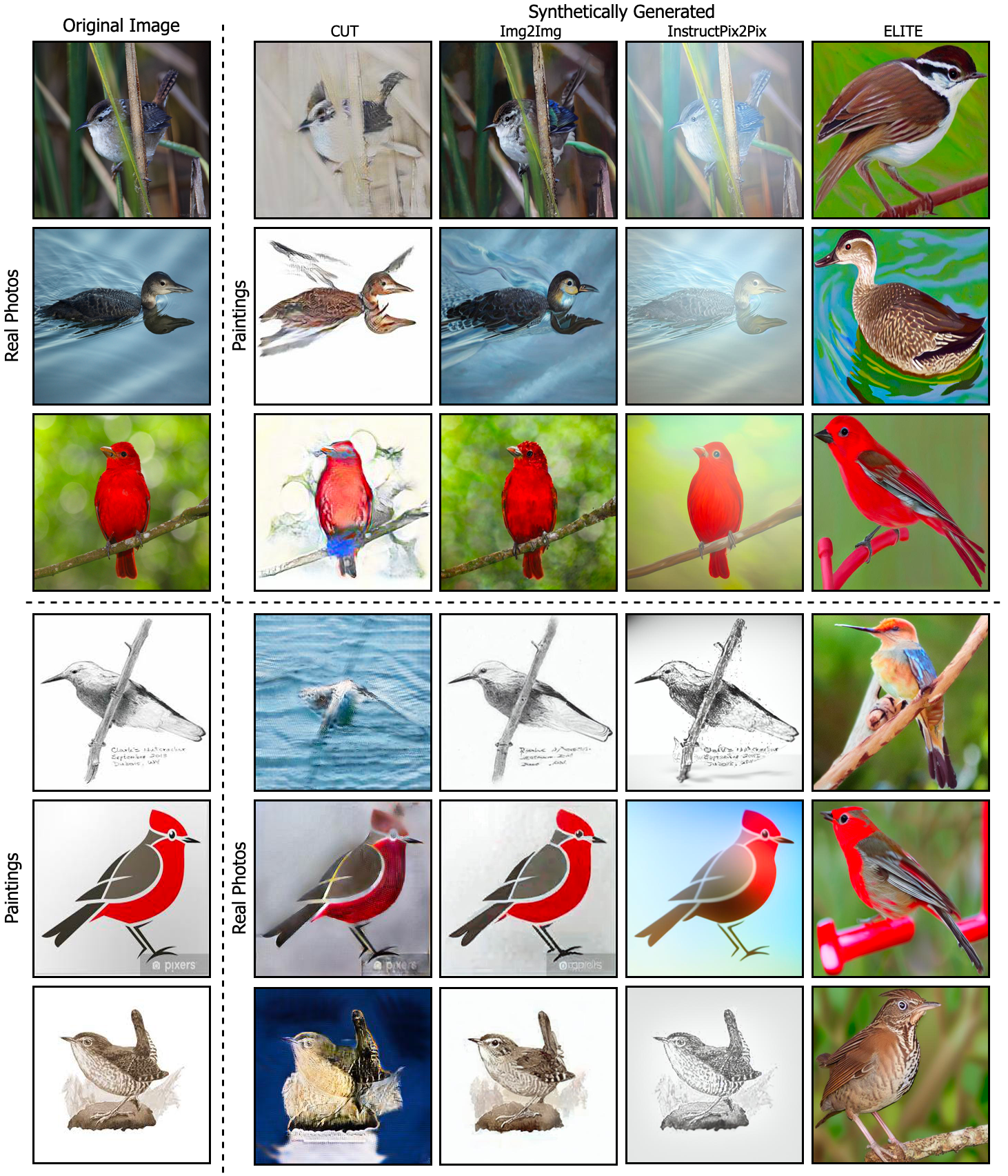}
    \caption{{\small \textbf{Synthetic examples from different translation methods for the CUB dataset.}}}
    \label{fig:cub_syn_examples} \vspace{-1mm}
\end{figure*}

\begin{figure*}[]
    \centering
    \includegraphics[width=0.95\linewidth,trim=0cm 0cm 0cm 0cm,clip]{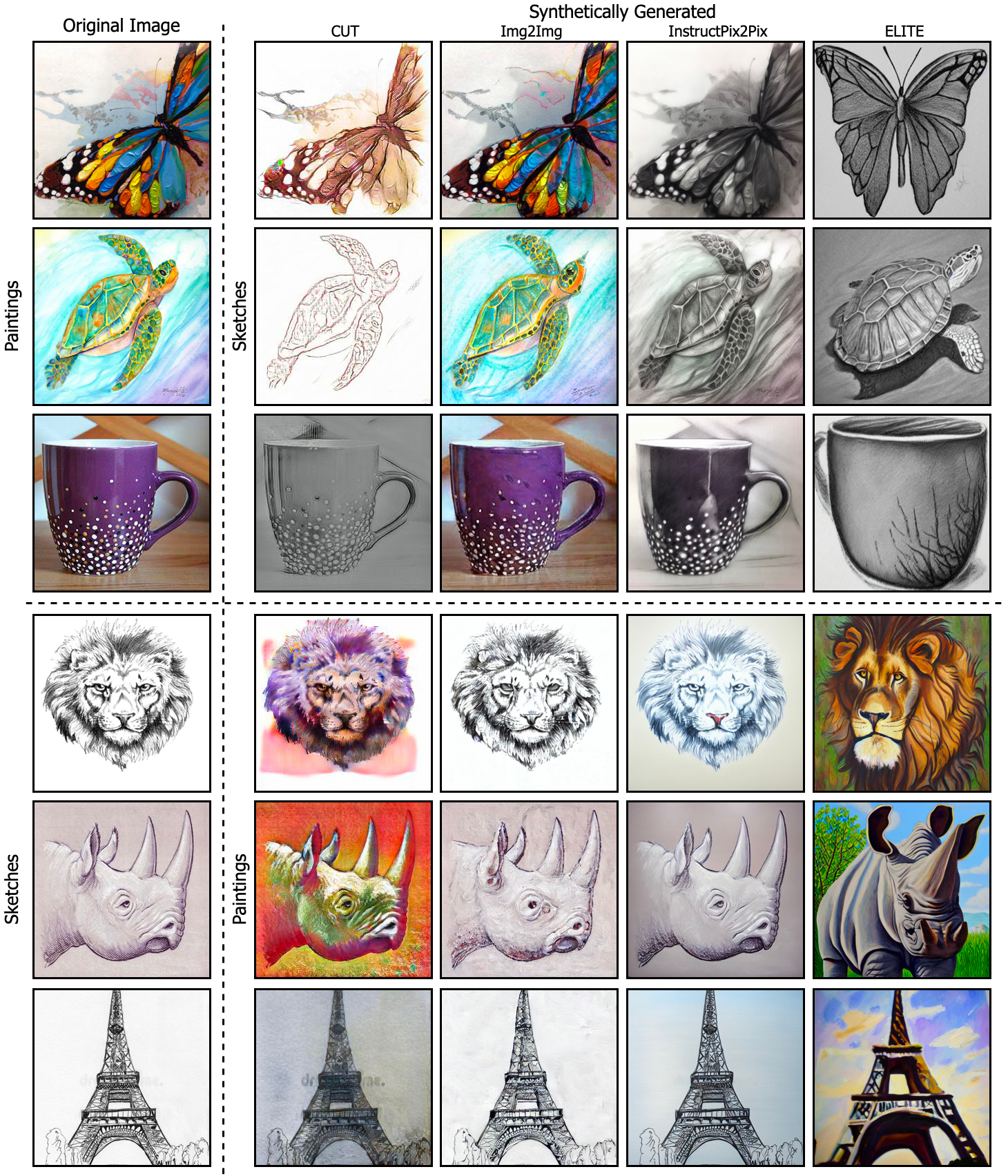}
    \caption{{\small \textbf{Synthetic examples from different translation methods for Paintings and Sketches in the DomainNet dataset.}}}
    \label{fig:domainnet_ps_syn_examples} \vspace{-1mm}
\end{figure*}

\begin{figure*}[]
    \centering
    \includegraphics[width=0.95\linewidth,trim=0cm 0cm 0cm 0cm,clip]{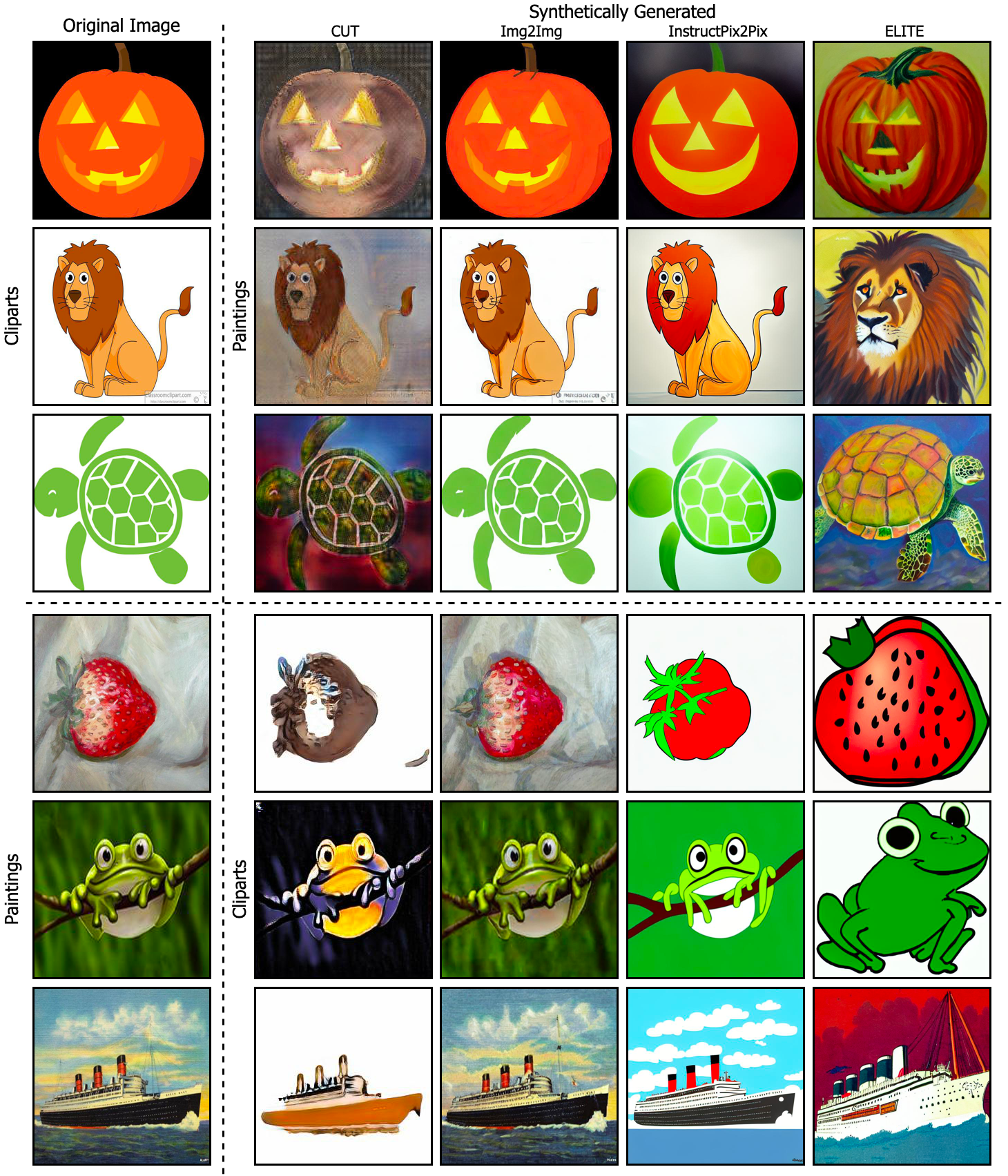}
    \caption{{\small \textbf{Synthetic examples from different translation methods for Cliparts and Paintings in the DomainNet dataset.}}}
    \label{fig:domainnet_cp_syn_examples} \vspace{-1mm}
\end{figure*}

\section{More Synthetic Generation Examples} \label{sec:more_qual_examples}
\cref{fig:cub_syn_examples,,fig:domainnet_ps_syn_examples,,fig:domainnet_cp_syn_examples} show some more examples of translations from the CUB and DomainNet datasets.

\end{document}